\documentclass[]{interact}

\usepackage{epstopdf}
\usepackage{amsmath}
\usepackage{amsthm}
\usepackage{amsfonts}
\usepackage{algorithm}
\usepackage{algpseudocode}
\usepackage{array}
\usepackage{url}
\usepackage{cite}

\usepackage{multirow}
\usepackage{graphicx}
\usepackage{subcaption}
\usepackage{tabularx}
\usepackage{listings}
\usepackage[table]{xcolor}

\theoremstyle{plain}

\theoremstyle{definition}

\theoremstyle{remark}
\newtheorem{remark}{Remark}

\begin{document}


\title{Adaptive Cluster-First Route-Second Decomposition for Industrial-Scale Vehicle Routing}

\author{
    Oguzhan Karaahmetoglu,
    Prof. Dr. Hyong Kim
    \thanks{O. Karaahmetoglu and Dr. Hyong Kim are with the Department of Electrical and Computer Engineering, Carnegie Mellon University, Pittsburgh, PA, 15213 USA e-mail: okaraahm@andrew.cmu.edu, kim@ece.cmu.edu}
}

\maketitle

\begin{abstract}
    Large-scale capacitated vehicle routing problems (CVRPs) are commonly addressed using cluster-first route-second (CFRS) approaches that split a routing instance into smaller, computationally tractable subproblems. Existing splitting methods typically rely on fixed partitioning rules, predefined optimization objectives, or learned policies, which may perform inconsistently across instances exhibiting different spatial, demand, and operational characteristics. In this work, we propose an adaptive CFRS system that formulates a decomposition procedure as an iterative decision-making process. Motivated by the recent success of large language models (LLMs) in reasoning and tool selection, the system employs an LLM as a high-level decision maker that analyzes the evolving decomposition state and selectively applies further clustering, balancing, and refinement operators. The proposed algorithm jointly partitions customers and vehicles, enabling capacity-aware clustering while adapting partitioning decisions to the characteristics of each problem. We evaluate the approach on synthetic and benchmark-derived CVRP instances containing up to 500,000 customers. Experimental results demonstrate competitive performance on benchmark-scale instances while exhibiting improved scalability and robust routing quality on substantially larger problems. These results highlight the potential of adaptive, LLM-guided decision support as a practical approach for industrial-scale vehicle routing and large-scale logistics planning.
\end{abstract}


\section{Introduction}
    \subsection{Preliminaries}
        Vehicle routing problems (VRPs) are a fundamental component of modern logistics and production systems, supporting transportation, distribution, and service operations for a wide range of industries \cite{toth2014vrp,laporte2009fifty}. As supply chains become increasingly dynamic and customer expectations continue to rise, organizations are required to solve routing problems of unprecedented scale while operating under diverse spatial, demand, fleet, and operational characteristics \cite{Mahvash19032017,Cortes03052022}. Such heterogeneity presents a significant challenge for CVRP methodologies, as decomposition and optimization strategies that perform well in one operating environment may generalize poorly to others \cite{lahyani2019hybrid, nikzad2023two}. Efficiently constructing high-quality routing plans under these varying conditions is essential for reducing routing costs, improving service quality, and supporting sustainable logistics operations \cite{bektas2011pollution,EdwinCheng17112022}. Consequently, the development of scalable and adaptive methodologies has become an increasingly important research direction for industrial decision support systems.
    
        Among the many variants of the VRP, the Capacitated Vehicle Routing Problem (CVRP) is one of the most widely studied and serves as a canonical formulation for numerous logistics and distribution applications \cite{toth2014vrp,laporte2009fifty}. In the CVRP, a fleet of capacity-constrained vehicles must serve a set of geographically distributed customers while minimizing routing cost. Despite its relatively simple formulation, the CVRP is NP-hard, and the computational complexity of obtaining high-quality solutions increases rapidly with problem size \cite{toth2014vrp,lenstra1981complexity}. Although substantial advances have been achieved through exact algorithms, metaheuristics, and hybrid optimization methods, efficiently solving industrial-scale instances involving tens or hundreds of thousands of customers while maintaining solution quality remains a significant computational challenge \cite{vidal2022hybrid}.
    
        To address the computational challenges of massive routing scenarios, many approaches adopt a cluster-first route-second (CFRS) paradigm, in which customers are partitioned into smaller subproblems that can be solved independently before being combined into a global solution \cite{toth2014vrp,alesiani2022constrained}. Existing partitioning methods can be broadly categorized as geometric, optimization-based, and learning-based approaches \cite{phiboonbanakit2018knowledge,li2022overview}. While these methods have shown success throughout a variety of settings, they generally rely on fixed partitioning rules, predefined optimization objectives, or learned policies trained on historical data \cite{alesiani2022constrained,gillett1974heuristic,christofides1969algorithm}. However, industrial environments often differ substantially in their spatial distributions, demand structure, fleet composition, and operational constraints, requiring partitioning strategies to be carefully designed for the characteristics of the routing instance \cite{lahyani2019hybrid,meyyappan2007real}. Hence, selecting an appropriate strategy remains a challenging decision-making task that directly influences downstream routing quality and computational efficiency \cite{alesiani2022constrained,vidal2022hybrid}.

        Motivated by these limitations, we propose an adaptive decomposition framework that employs a large language model (LLM) as a high-level decision-making component for routing decomposition. Recent advances have revealed the ability of LLMs to perform complex reasoning and coordinate external tools, making them well suited for sequential decision-support tasks beyond natural language processing \cite{jannelli2026agentic,li2026llm,yao2023react}. The LLM iteratively analyzes the incumbent decomposition plan, identifies potential inefficiencies, and selects refinement actions based on the observed state without directly assigning individual customers to clusters. These actions are executed through a collection of clustering, balancing, and partition-modification heuristics exposed as optimization tools, allowing the methodology to leverage established decomposition methods while adaptively selecting strategies appropriate for the distribution of the CVRP dataset. Through this iterative process, our approach continuously refines the partitions according to the spatial, demand, fleet, and operational configuration of the problem, as opposed to relying on a single predefined clustering objective or fixed partitioning rule.

    \subsection{Prior Art}
        Decomposition has long been recognized as an effective strategy for improving the computational tractability of large-scale VRPs by dividing an optimization formulation into smaller, more manageable subproblems \cite{laporte2009fifty,toth2014vrp}. Instead of solving the original task as a single monolithic process, decomposition approaches partition customers and resources into subproblems that can be solved independently before being integrated into a global operational plan \cite{santini2023decomposition}. Among the various strategies proposed in the literature, clustering-based approaches have become particularly popular due to their simplicity, scalability, and compatibility with a broad range of algorithms \cite{gillett1974heuristic,christofides1969algorithm,alesiani2022constrained}.
    
        Geometric partitioning methods decompose instances primarily according to the spatial distribution of customers. Representative examples include sweep-based clustering, grid-based partitioning, and recursive spatial splitting \cite{gillett1974heuristic,christofides1969algorithm,bent2008spatial}. These approaches are attractive due to their simplicity, computational efficiency, and scalability to very large VRP instances. However, partitioning decisions are primarily driven by geographic information and generally do not explicitly account for customer demand, vehicle capacities, or other operational considerations \cite{alesiani2022constrained,bramel1997logic}. Thus, additional balancing or repair procedures are often required to improve workload distribution and restore routing feasibility.
    
        Optimization-based clustering methods extend geometric partitioning by incorporating operational objectives directly into the decomposition process. In addition to spatial compactness, these approaches explicitly consider customer demand, vehicle capacities, workload balancing, and other resource-related constraints when constructing clusters \cite{alesiani2022constrained,abdellaoui2024towards}. By integrating these considerations into the process, optimization-based methods generally produce partitions that better reflect routing requirements and available resources. Recent work has further explored optimization-driven strategies for large instances, demonstrating strong scalability while maintaining high-quality logistics solutions \cite{santini2023decomposition,tu2017spatial}. Therefore, optimization-based clustering has become a widely adopted approach for computationally demanding instances.
    
        Despite these advantages, optimization-based and learning-based approaches remain dependent on the objectives specified during partition construction or model training. Common optimization objectives include balancing customer counts \cite{calvete2025balancing}, total demand \cite{jingjing2022cluster}, vehicle capacity \cite{linfati2022mathematical}, and spatial compactness for all clusters \cite{abdellaoui2024towards,linfati2022mathematical}, while learning-based methods are typically trained to optimize surrogate objectives derived from historical observations encountered \cite{chin2026neural,li2022overview}. Since each instance can vary substantially in their spatial structure, demand distribution, fleet composition, and operational constraints \cite{uchoa2017new}, the relative importance of these objectives may also vary across operating environments. Consequently, designing effective decomposition strategies requires balancing multiple, and often competing, routing considerations, motivating adaptive approaches capable of selecting decomposition actions according to the statistical properties of the data.
    
        Recent advances in large language models (LLMs) have generated growing interest in their application to combinatorial optimization and industrial decision-support systems \cite{da2026large,xiu2026large}. Beyond direct solution generation, LLMs have shown the ability to reason over complex system states, coordinate external optimization tools, and sequentially select actions within iterative workflows \cite{thind2025optimai,yao2023react}. These capabilities make them well suited for optimization formulations in which effective decisions depend on the characteristics of the evolving solution rather than a single predefined objective. This perspective has recently led to the emergence of agentic optimization frameworks, where LLMs guide optimization by repeatedly analyzing intermediate states, selecting appropriate actions, and incorporating feedback from specialized algorithms and heuristics \cite{xi2025rise, li2026llm}. Such developments suggest that LLMs can serve as adaptive decision-making components within decomposition workflows while leveraging established optimization techniques for execution.

    \subsection{Contributions}
        In this paper, we make the following contributions.
        \begin{itemize}
            \item We propose an adaptive cluster-first route-second framework for large-scale CVRPs that formulates decomposition as an iterative decision-making process. The approach employs an LLM to analyze intermediate partitions and coordinate clustering, balancing, and refinement operations through a collection of optimization tools.
        
            \item We introduce a joint customer-vehicle splitting strategy that explicitly incorporates resource availability into the decomposition process, enabling capacity-aware partitions for heterogeneous fleets.
            
            \item We develop a large-scale evaluation benchmark spanning both synthetic and benchmark-derived CVRP instances, including VRP data containing up to 500,000 customers under diverse spatial, demand, and operational settings.
            
            \item We demonstrate that our approach yields high performance on benchmark-scale instances while exhibiting improved scalability and robust performance on substantially larger routing scenarios.
        \end{itemize}

\section{Problem Formulation}
    In this work, we consider the CVRP with heterogeneous vehicle capacities and unsplittable customer demands (i.e., no more than one vehicle can visit the same customer). Given a set of customers $ N = \{i\}_{i=1}^{|N|} $, associated demands $ f_i, i\in N $, and a fleet of vehicles $ K $ with limited capacities $ F_k, k\in K $, the objective is to construct a set of routes that minimizes the total routing cost while respecting vehicle capacity constraints. Although the primary goal is to serve all customer demand, the formulation permits unserved customers through a demand-weighted penalty term $ \lambda \in \mathbb{R}^+ $ in the objective, thereby discouraging missed demand while allowing infeasible demand levels or resource shortages to be represented within the model.

    \begin{table}[h]
        \caption{CVRP Formulation Notation}
        \label{tab:notation}
        \centering
        \begin{tabularx}{\columnwidth}{lX}
        \hline
        Symbol & Description \\
        \hline
        $N$ & Set of customers \\
        $K$ & Set of vehicles \\
        $V=\{0\}\cup N$ & Set of all nodes including the depot node $0$ \\
        $f_i$ & Demand of customer $i$ \\
        $F_k$ & Capacity of vehicle $k$ \\
        $q_{ijk}$ & Travel cost from node $i$ to node $j$ using vehicle $k$ \\
        $\lambda$ & Penalty coefficient for unserved demand \\
        $a_{ijk}$ & Binary variable indicating whether vehicle $k$ traverses arc $(i,j)$ \\
        $b_{ik}$ & Binary variable indicating whether customer $i$ is assigned to vehicle $k$ \\
        $u_i$ & Binary variable indicating whether customer $i$ is left unserved \\
        \hline
        \end{tabularx}
    \end{table}

\begin{figure}[t]
        \centering
        \begin{subfigure}[t]{0.321\textwidth}
            \centering
            \includegraphics[width=\textwidth]{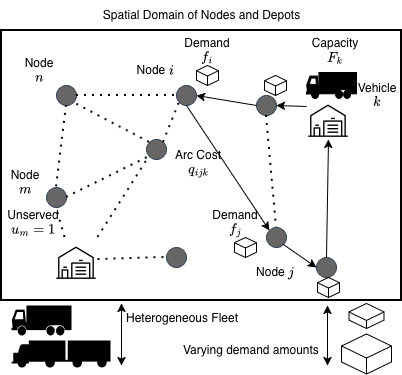}
            \caption{Heterogeneous Fleet CVRP. Customers with unsplittable demands are assigned to vehicles with heterogeneous capacities while minimizing routing cost and penalizing unserved demand.}
            \label{fig:hcvrp_problem}
        \end{subfigure}
        \hfill
        \begin{subfigure}[t]{0.321\textwidth}
            \centering
            \includegraphics[width=\textwidth]{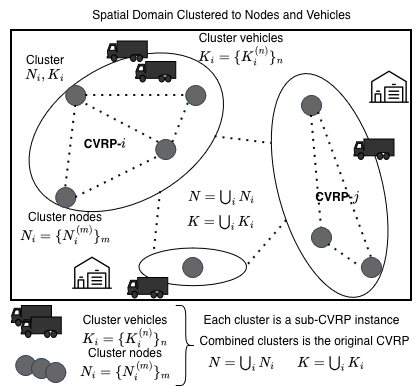}
            \caption{Cluster-First Route-Second (CFRS). The problem is decomposed into customer-vehicle subproblems that are solved independently and later refined through collaboration and global optimization.}
            \label{fig:cfrs_paradigm}
        \end{subfigure}
        \hfill
        \begin{subfigure}[t]{0.321\textwidth}
            \centering
            \includegraphics[width=\textwidth]{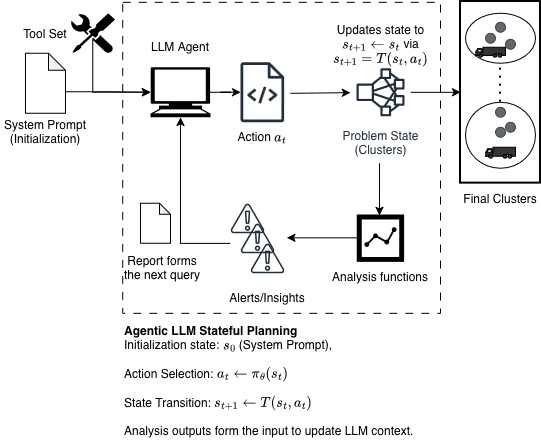}
            \caption{LLM-guided planning. An LLM-guided decision maker analyzes intermediate solutions, selects actions, invokes helper tools, and iteratively updates the output.}
            \label{fig:agentic_planning}
        \end{subfigure}
    
        \caption{
        Overview of the optimization setting and solution method. Figure~\ref{fig:hcvrp_problem} illustrates the heterogeneous fleet CVRP considered in this work. Figure~\ref{fig:cfrs_paradigm} presents the Cluster-First Route-Second paradigm used to scale routing optimization to very large instances. Figure~\ref{fig:agentic_planning} shows how agentic AI can be incorporated into planning workflows by coupling language-model reasoning with optimization and analysis tools.
        }
        \label{fig:overview}
    \end{figure}

    Using the notation introduced in Table~\ref{tab:notation}, we illustrate a generic CVRP instance with the described requirements in Figure~\ref{fig:hcvrp_problem}. Following standard CVRP formulations \cite{toth2014vrp,laporte2009fifty}, the formulation can be expressed as the following integer linear program:
    \begin{align}
    \label{eqn:formulation}
        \min_{a,b,u}\quad &
        \sum_{k\in K}\sum_{i\in V}\sum_{j\in V} q_{ijk}a_{ijk} + \lambda\sum_{i\in N} f_i u_i\\
        \text{s.t.}\quad
        &
        \sum_{k\in K} b_{ik}+u_i=1 && \forall i\in N \\
        &
        \sum_{j\in V} a_{ijk}=b_{ik} && \forall i\in N,\forall k\in K \\
        &
        \sum_{j\in V} a_{jik}=b_{ik} && \forall i\in N,\forall k\in K \\
        &
        \sum_{i\in N} f_i b_{ik} \le F_k && \forall k\in K \\
        &
        a_{ijk}\in\{0,1\} && \forall i,j\in V,\forall k\in K \\
        &
        b_{ik}\in\{0,1\} && \forall i\in N,\forall k\in K \\
        &u_i\in\{0,1\} && \forall i\in N.
    \end{align}
    Subtour elimination constraints are omitted for brevity but are assumed to be present to prevent disconnected cycles that do not include the depot and to ensure that each vehicle route forms a connected tour.

    The objective in \eqref{eqn:formulation} minimizes total delivery cost while penalizing unserved demand through the parameter $\lambda$, thereby discouraging customer omissions whenever feasible. Constraints enforce customer assignment, route continuity, and vehicle capacity limits, while ensuring that each customer is either assigned to a vehicle or marked as unserved.

    A solution is defined as a collection of routes $S=\{R_1,\ldots,R_m\}$, where each route $ R_k=(0,R_k^{(1)},\ldots,R_k^{(r)},0), R_k^{(i)}\in N $ is an ordered sequence of customer visits beginning and ending at the depot. The cost of a route is defined as $C(R_k)=\sum_{t=0}^{r} q_{R_k^{(t)},R_k^{(t+1)},k}$, and the total solution cost is $C(S)=\sum_{R_k\in S} C(R_k)$. Let $ U(S)\subseteq N $ denote the set of customers not served by any route in $S$. The unserved demand of a solution is defined as $ M(S)=\sum_{i\in U(S)} f_i $.

    In the next subsections, we explain how decomposition in CVRP is formulated using CFRS paradigm and the agentic systems are used in solving combinatorial optimization problems.

\subsection{CFRS Paradigm Formulation}
    Given a CVRP instance with customers and vehicles $(N,K)$, a CFRS method decomposes the original planning task into a collection of smaller subproblems as shown in Figure \ref{fig:cfrs_paradigm}. Let $\mathcal{P}=\{(N_1, K_1),\ldots,(N_p,K_p)\}$ denote a partition of the customer and vehicle sets, where $N_c\subseteq N$ and $K_c\subseteq K$ represent the customers and vehicles assigned to partition $c$. The decomposition procedure is treated as a black-box process that generates the partitions according to a particular clustering strategy such as Sweep \cite{gillett1974heuristic}.

    For each partition $(N_c,K_c)$, an independent CVRP subproblem is constructed using the formulation presented in the previous section, restricted to the customers and vehicles assigned to that partition. Solving the subproblem yields a partial solution $ S_c=\{R_k\}_{k\in K_c} $, where unused vehicles correspond to empty routes.

    The final solution is obtained by combining the routes produced by all partitions, $S=\bigcup_{c=1}^{p} S_c$. Since each subproblem is solved independently, partitioning decisions directly affect the resulting solution. For example, if a cluster has lower total capacity than total demand $\sum_{k\in K_c}F_k < \sum_{i\in N_c}f_i$, then partition $c$ cannot serve all assigned demand, leading to unserved customers in the final solution. Likewise, restricting customers and vehicles to a particular partition may increase routing cost even when each subproblem is solved optimally.

    Abstracting away the details of a particular clustering algorithm, a decomposition method can be represented as an operator $D_\theta:(N,K)\rightarrow\mathcal{P}$, where $\theta$ denotes the parameters governing the decomposition policy. Different CFRS approaches correspond to different implementations of the operator.

    \begin{remark}
        The operator $D_\theta$ need not jointly partition customers and vehicles. Many CFRS methods instead employ separate operators $D_\theta^N(N)$ and $D_\theta^K(K)$ to generate customer and vehicle partitions independently. For example, customers may be clustered using k-means \cite{lloyd1982least} while vehicles are assigned with equal counts.
    \end{remark}

    Within this framework, $\theta$ denotes the information that determines how a partition is generated. For optimization-based methods, $\theta$ typically includes the clustering objective, balancing criteria, and algorithmic settings. For learning-based methods, $\theta$ corresponds to the parameters of the trained model. Thus, many existing CFRS approaches can be viewed as instances of the same operator $D_\theta$, differing only in how partitioning decisions are produced.

    Certain CFRS approaches additionally employ a global refinement phase after the partition solutions have been combined into the solution $S=\bigcup_{c=1}^{p}S_c$ \cite{kerscher2024spatial}. This phase applies a post-processing operator $R(S)$ that modifies routes across partition boundaries in order to reduce delivery cost, recover unserved demand, or both. 
    
    In the following subsection, we explain how we use agentic LLMs to build hierarchical decompositions iteratively.

\subsection{Agentic AI in Adaptive Planning and Optimization}
    Agentic optimization systems treat optimization as a sequential decision-making process as in Figure \ref{fig:agentic_planning}. At each iteration, an agent observes the state of the optimization procedure, selects an action, and updates the state based on the outcome of that action. Such a process can be represented by the tuple $(\mathcal{S},\mathcal{A},\pi_\theta,T)$, where $\mathcal{S}$ denotes the state space and $\mathcal{A}$ denotes the set of available actions. The policy and state transition operators are defined as
    \begin{align}
        \pi_\theta &: \mathcal{S}\rightarrow\mathcal{A}, \\
        T &: \mathcal{S}\times\mathcal{A}\rightarrow\mathcal{S},
    \end{align}
    where $\pi_\theta$ denotes a policy parameterized by $\theta$ and $T$ denotes the state transition operator.

    The policy parameters $\theta$ define the decision-making mechanism used to select actions from the action space $\mathcal{A}$, which corresponds to the set of tools available to the agent. In CFRS settings, these tools may include clustering heuristics, exact algorithms, analysis procedures, or local search operators. Different agentic optimization approaches can therefore be viewed as different instantiations of $(\mathcal{S},\mathcal{A},\pi_\theta,T)$, differing in their state representation, available tools, and decision policy.

    We present our methodology and how it connects to this formulation in the next section.

\section{Methodology}
\begin{figure}[t]
        \centering
        \begin{subfigure}[b]{0.325\textwidth}
            \centering
            \includegraphics[width=\linewidth]{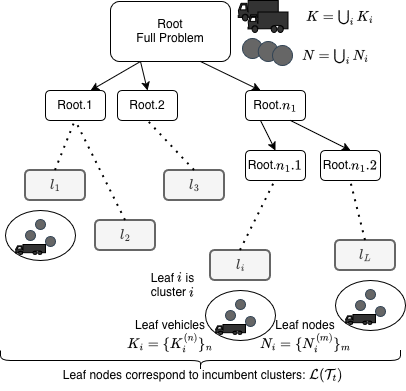}
            \caption{Hierarchically decomposed CVRP, where the leaf nodes correspond to the clusters.}
            \label{fig:tree-clusters}
        \end{subfigure}
        \hfill
        \begin{subfigure}[b]{0.325\textwidth}
            \centering
            \includegraphics[width=\linewidth]{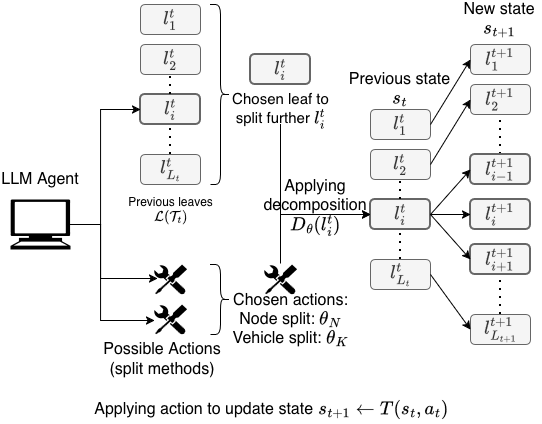}
            \caption{Decomposition agent performing a state transitions by selecting a leaf node, customer and vehicle split tools, refining the selected node further.}
            \label{fig:state-transitions-decomp}
        \end{subfigure}
        \hfill
        \begin{subfigure}[b]{0.325\textwidth}
            \centering
            \includegraphics[width=\linewidth]{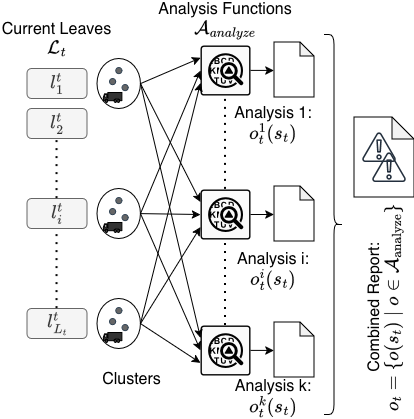}
            \caption{Leaf nodes are passed to every analysis tools, which produce reports signaling possible issues (e.g., no vehicles in cluster). Reports are combined and passed to the agent.}
            \label{fig:agentic-action-selection}
        \end{subfigure}
    
        \caption{LLM-guided decomposition. The planner maintains a hierarchical decomposition state, applies tool-based actions to selected leaves, and uses alerts from the incumbent partitions to guide subsequent decisions.}
        \label{fig:agentic_framework}
    \end{figure}

    The proposed framework is illustrated in Figure~\ref{fig:agentic_framework}. The following sections introduce the cluster-tree representation, the associated state-transition model, and the agent decision process used to guide decomposition and refinement decisions.

\subsection{Cluster Tree Representation}
    The decomposition is represented by a rooted tree $\mathcal{T}=(\mathcal{N},\mathcal{E})$, where each node $v\in\mathcal{N}$ corresponds to a subproblem as illustrated in Figure~\ref{fig:tree-clusters}. A node is defined by the tuple $ v=(N_v,K_v) $, where $N_v\subseteq N$ denotes the customers assigned to the node and $K_v\subseteq K$ denotes the vehicles assigned to the node.


    The root node corresponds to the original CVRP instance and is defined as $(N,K)$. Given a node $v=(N_v,K_v)$, a decomposition operation produces a set of child nodes $\mathcal{C}(v)=\{v_1,\ldots,v_m\}$ such that
    \begin{equation}
        N_v=\bigcup_{u\in\mathcal{C}(v)} N_u,
        \qquad
        N_i\cap N_j=\emptyset
        \quad \forall i\neq j.
    \end{equation}
    Vehicle assignments are similarly distributed among the child nodes according to the chosen operator. The set of leaf nodes for a given tree $ \mathcal{T} $ is denoted by $\mathcal{L}(\mathcal{T})=\{l_1,\ldots,l_m\}$, where each leaf represents a cluster that can be independently solved as a smaller CVRP subproblem. Collectively, the leaf nodes define the final decomposition
    \begin{equation}
        \mathcal{P}(\mathcal{T}) = \{(N_l,K_l)\mid l\in\mathcal{L}(\mathcal{T})\}.
    \end{equation}

    The cluster tree provides a hierarchical representation of the incumbent partitions. The following subsection formalizes this hierarchical decomposition as a state-transition process and introduces the transition operators that modify the tree during the iteration.

\subsection{State and Transitions in Hierarchical Decomposition}
    The iterative process is modeled as a sequence of state transitions over the cluster tree as illustrated in Figure~\ref{fig:state-transitions-decomp}. Internal nodes represent intermediate partitions, whereas leaf nodes correspond to the active clusters. Therefore, the state at iteration $t$ is defined by the leaf set of the evolving tree, $s_t=\mathcal{L}(\mathcal{T}_t) = \{l_i^{(t)}\}_i$, which induces the decomposition $\mathcal{P}(s_t)=\{(N_n,K_n)\mid n\in\mathcal{L}(\mathcal{T}_t)\}$.

    At each iteration of state transitions, an operator $D_{\theta_t}$ is applied to the state, producing a new state according to $s_{t+1}=D_{\theta_t}(s_t)$. The parameters $\theta_t$ determine the behavior of the operator and correspond to agent-selected actions. Repeated application of decomposition operators generates a sequence of states $s_0\rightarrow s_1\rightarrow\cdots\rightarrow s_T$, where the final state $s_T$ determines the partition set used by the CFRS framework.

    State transitions are induced by actions selected from the action space $\mathcal{A}$. Each action identifies a leaf within the cluster tree together with an operation to be applied to that node. Executing the action modifies the tree and consequently produces a new state. The action space consists of three categories of transition operations:
    \begin{itemize}
        \item \textbf{Split}: Decomposes one or more leaf nodes into smaller subproblems.
        \item \textbf{Redo}: Reverts a previous operator and applies an alternative strategy.
        \item \textbf{Stop}: Terminates the iterations and returns the current partition set.
    \end{itemize}

    A partition action applies customer and vehicle decomposition procedures to a leaf node $v=(N_v,K_v)$. Let $\theta=(\theta_N,\theta_K,\alpha)$ denote the decomposition parameters, where $\theta_N$ and $\theta_K$ specify the customer and vehicle clustering tools, respectively, and $\alpha$ controls the split granularity. The partition operator generates a set of child nodes $D^{\mathrm{part}}_{\theta}(v)$ that replace $v$ in the leaf set, thereby producing a new state (and new set of partitions).
    
    A repartition action revisits a previously expanded node $v=(N_v,K_v)$ using an alternative parameter set $\theta'=(\theta'_N,\theta'_K,\alpha')$. The descendants of $v$ are removed, $v$ is restored as a leaf node, and a new partition operation is applied. This enables the process to revise earlier decisions without reconstructing the entire tree.
    
    Finally, a stop action terminates the algorithm and returns the partition set induced by the leaf nodes of the cluster tree. This action may be selected when the incumbent has reached the desired level of granularity, available computational resources have been exhausted, or further partitioning is unlikely to improve the quality of the result.

    The actions defined above specify how the state may be modified, while the particular clustering procedures are determined by the selected parameters $\theta_N$ and $\theta_K$. The complete set of operations given as agent tools is provided in Appendix~\ref{appendix:split-tool}-\ref{appendix:stop-tool}.

\subsection{Action Selection and State Description}
    As illustrated in Figure~\ref{fig:agentic-action-selection}, each decomposition state is first evaluated using a collection of analysis tools $\mathcal{A}_{\mathrm{analyze}}$. Applied to the leaf nodes of the incumbent, these tools generate reports describing properties such as cluster compactness, demand-capacity balance, and resource utilization. The resulting reports are aggregated into an observation
    \begin{equation}
        o_t=\Phi\!\left(\{a(l^{(t)})\mid l^{(t)}\in s_t,\; a\in\mathcal{A}_{\mathrm{analyze}}\}\right),
    \end{equation}
    where $\Phi$ denotes the aggregation procedure.
    
    The observation $o_t$ together with the available transition actions $\mathcal{A}_{\mathrm{trans}}$ is provided to the agent, which selects an action and its parameters according to
    \begin{equation}
        (\hat{a}_t,\hat{\theta}_t)=\pi_\theta(o_t,\mathcal{A}_{\mathrm{trans}}),
    \end{equation}
    where $\hat{a}_t\in\mathcal{A}_{\mathrm{trans}}$ denotes the selected action and $\hat{\theta}_t$ denotes the corresponding parameters.
    
    The analysis reports contain global statistics and partition-level alerts that identify potential deficiencies such as oversized partitions, insufficient capacity, or imbalanced vehicle allocation. The complete set of analysis signals is provided in Appendix~\ref{appendix:analysis-tools}. Since these reports are incorporated directly into the agent context, they are compressed and restricted by an analysis limit parameter $A_{\max}$, which returns only the most relevant observations.

    The selected action is subsequently validated against the latest partitions. If the action is applicable, the corresponding state transition
    \begin{equation}
        s_{t+1}=T(s_t,\hat{a}_t,\hat{\theta}_t)
    \end{equation}
    is executed. Otherwise, a warning message $w_t$ describing the violation is appended to the agent context, yielding the updated observation
    \begin{equation}
        o_t' = o_t \cup \{w_t\}.
    \end{equation}
    The agent then performs a new action-selection step using $o_t'$. This process repeats until a stop action is selected, yielding the final decomposition state and corresponding partition set. In the next subsection, we give this loop in a structured manner.

    We initialize the agent with the system prompt provided in Appendix~\ref{appendix:sysprompt} along with the initial set of analysis report that gives necessary context to the LLM agent. In the next subsection, we go through the entire pipeline starting from this initializing context to reach to a decomposition plan.

\subsection{Adaptive Decomposition Full Pipeline}
    The overall mechanism is summarized in Algorithm~\ref{alg:agentic_decomposition}. Starting from the initial state $s_0=\{\mathrm{Root}\}$, the framework iteratively generates observations, selects a transition action, and applies the corresponding state transition. The process continues until a stop action is selected, yielding a final state whose leaf nodes define the partition set.

    \begin{algorithm}[H]
    \caption{Agentic Hierarchical Decomposition}
    \label{alg:agentic_decomposition}
    \begin{algorithmic}[1]
    \State Initialize cluster tree $\mathcal{T}$ with root node $(N,K)$
    \State Initialize decomposition state $s_0=\{\mathrm{Root}\}$
    \While{stop action not selected}
        \State Generate observation $o_t$ using analysis tools $\mathcal{A}_{\mathrm{analyze}}$
        \State Select action and parameters $(\hat{a}_t,\hat{\theta}_t)=\pi_\theta(o_t,\mathcal{A}_{\mathrm{trans}})$
        \If{$(\hat{a}_t,\hat{\theta}_t)$ is valid}
            \State Apply transition $s_{t+1}=T(s_t,\hat{a}_t,\hat{\theta}_t)$
        \Else
            \State Generate warning $w_t$
            \State Update observation $o_t \leftarrow o_t \cup \{w_t\}$
        \EndIf
    \EndWhile
    \State Return partition set $\mathcal{P}(s_t)$
    \end{algorithmic}
    \end{algorithm}

    The cluster action is implemented through a compound tool that applies one customer clustering method and one vehicle distribution method to a selected leaf cluster. In addition to the tools, the action specifies a level that controls the granularity of the resulting child clusters. The procedure is summarized in Algorithm~\ref{alg:split_partition}.
    
    \begin{algorithm}[H]
        \caption{Decomposition Tool}
        \label{alg:split_partition}
        \begin{algorithmic}[1]
        \State Read target partition $n=(N_n,K_n)$
        \State Reject if $n$ is not a leaf node
        \State Reject if maximum tree depth reached
        \State Compute num. children partitions
        \State Apply selected customer split
        \State Apply selected vehicle split
        \State Return status message as success/fail
        \end{algorithmic}
    \end{algorithm}
    
    The number of child clusters is computed dynamically rather than fixed in advance. Let $n$ denote the number of customers in the selected partition, $N_{\min}$ the minimum partition size, $d$ the tree depth, and $D_{\max}$ the maximum tree depth. The framework first estimates the number of routing-ready leaf partitions required by the chosen partition and distributes this estimate across the remaining tree depth to obtain a baseline branching factor. The decomposition level then adjusts this branching factor according to
    \begin{equation}
        k=
        \begin{cases}
            \max(2,b-1), & \text{low},\\
            b, & \text{medium},\\
            b+1, & \text{high},\\
            \left\lceil \frac{n}{N_{\min}} \right\rceil, & \text{finalized},
        \end{cases}
    \end{equation}
    where $b$ denotes the baseline branching factor computed from the estimated number of remaining leaf partitions and the remaining tree depth. The resulting value is further constrained to ensure that no child partition violates the minimum partition size requirement. This approach allows the same tool to adapt to partitions of different sizes and depths while remaining consistent with both local partition quality and global decomposition objectives.

\subsection{Algorithm Parameters}
    The configurable parameters of the framework are summarized in Table~\ref{tab:framework_parameters}. Parameters $P_{\max}$, $T_{\max}$, and $C_{\max}$ primarily govern the termination behavior of Algorithm~\ref{alg:agentic_decomposition}, while $N_{\min}$ and $L_{\max}$ regulate the granularity of the resulting decomposition by constraining admissible state transitions. The parameter $A_{\max}$ limits the amount of diagnostic information provided to the agent, and $\pi_\theta$ together with $\tau$ determine the action-selection policy. We evaluate two lightweight open-weight language models, Qwen2.5 and Qwen3, which provide an attractive balance between reasoning capability and computational efficiency, making them suitable candidates for evaluating the proposed agentic decomposition framework.

    \begin{table}[t]
        \caption{Our Parameters.}
        \label{tab:framework_parameters}
        \centering
        \begin{tabularx}{\columnwidth}{lXl}
        \hline
        \textbf{Symbol} & \textbf{Parameter} & \textbf{Choice} \\
        \hline
        $P_{\max}$ & Target number of clusters & 25 \\
        $\pi_\theta$ & LLM policy & Qwen2.5 \cite{qwen2025qwen25technicalreport} / Qwen3 \cite{yang2025qwen3technicalreport} \\
        $T_{\max}$ & Maximum runtime & 300 s \\
        $C_{\max}$ & Maximum agent iterations & 30 \\
        $N_{\min}$ & Minimum cluster size & 2,000 \\
        $\tau$ & Temperature & 0.9 \\
        $A_{\max}$ & Analysis limit & 10 \\
        $L_{\max}$ & Maximum tree depth & 4 \\
        \hline
        \end{tabularx}
    \end{table}

    The parameter values reported in Table~\ref{tab:framework_parameters} were selected through preliminary tuning on synthetically generated CVRP instances containing up to 500,000 customers, using customer miss rate and clustering time as the primary selection criteria. Synthetic instances are randomly generated and we separate the tuning set from the evaluation set for unbiased choice of parameters. Details of the synthetic instance generator are provided in experimental setup.

\section{Experimental Setup}
    In this section, we describe the experiment setup and present the performance of our method. We first present the comparison models used in our benchmark, followed by the metrics we use to assess the comparison. Finally, we present the results obtained on synthetic and public data instances.

\subsection{Comparison Models}
    The evaluated methods span geometric, optimization-based, learning-based, and adaptive approaches. Geometric baselines include \textbf{RND} (Random), \textbf{SWP} (Sweep) \cite{gillett1974heuristic}, \textbf{GRD} (Grid), and \textbf{QDR} (Quadrant), which partition customers according to spatial proximity. Optimization-based methods include \textbf{BKM} (Balanced K-Means), which balances cluster compactness and assigned vehicle capacity, \textbf{ENT} (Entropy), which promotes balanced customer assignments by maximizing partition entropy, \textbf{CCBC} (Constrained Centroid-Based Clusterin) \cite{abdellaoui2024towards} and \textbf{CC-CVRP} (Capacity Constrainted CVRP) \cite{alesiani2022constrained}, which explicitly incorporates vehicle capacity constraints during partition construction. Learning-based methods include \textbf{GLOP} \cite{ye2024glop}, which learns policies from data. Finally, \textbf{A-DEC} denotes the proposed adaptive decomposition framework, which iteratively refines a hierarchical cluster tree through tool-guided decision making.

\subsection{Evaluation and Analysis Metrics}
    The objective of the experimental evaluation is twofold. First, we assess whether the proposed procedure enables the construction of high-quality solutions with low customer miss rates. Second, we evaluate its robustness under varying scales and operating conditions representative of industrial environments. To this end, all methods are evaluated using a common downstream procedure. First, the decomposition method generates a set of clusters, which are then solved independently using OR-Tools Guided Local Search (GLS) \cite{ortools_routing}. Cluster-level routing instances are solved in parallel for up to 300 seconds or until all assigned demand has been served. The resulting partial solutions are subsequently merged into a single solution and provided as an initial solution to a global OR-Tools GLS solver. The global solver is executed for up to 30 minutes, allowing route exchanges and other long-range modifications within cluster boundaries. Performance metrics are reported on the resulting merged solution.

    The first metric is the decomposition runtime, denoted by $T_{\mathrm{cluster}}$, which measures the wall-clock time required to generate the resulting partitions. This metric reflects the computational overhead introduced by the clustering stage and serves as an indicator of practical scalability.

    Customer miss rate is measured on the final routing solution after global refinement. Let $d_i$ denote the demand of customer $i$ and let $u_i\in\{0,1\}$ indicate whether customer $i$ remains unserved in the final solution. The miss rate is computed as
    \begin{equation}
        M=\frac{\sum_{i\in N} d_i u_i}{\sum_{i\in N} d_i}.
    \end{equation}

    Final distance is measured on the globally refined solution and defined as
    \begin{equation}
        D=\sum_{r\in R}\sum_{(i,j)\in r} c_{ij},
        \label{eqn:dist}
    \end{equation}
    where $R$ denotes the set of routes in the final solution and $c_{ij}$ denotes the travel cost between locations $i$ and $j$, computed as the Euclidean distance
    \[
    c_{ij}=\sqrt{(x_i-x_j)^2+(y_i-y_j)^2}.
    \]
    In the results, we interchangeably use the terms \emph{distance} and \emph{cost} to refer to the metric defined in \eqref{eqn:dist}.

    From an operational perspective, these metrics capture complementary aspects of routing quality. Decomposition runtime reflects the computational overhead introduced by the decision-support process, while customer miss rate and distance quantify the quality of the resulting operational plan. Together, they measure the trade-off between responsiveness and solution quality that is critical in logistics planning involving hundreds of thousands of customers.

    We focus exclusively on downstream metrics because the objective of a decomposition strategy is not to optimize partition structure in isolation, but to support effective operational decision making through high-quality plans. In practical logistics systems, decomposition serves as an intermediate decision process whose value is ultimately determined by its impact on customer service levels, routing efficiency, and computational responsiveness. If a benchmark method directly produces routes without exposing intermediate partitions (such as \textbf{GLOP}), partition-specific metrics are reported as $-$. Methods that fail to produce a valid solution within the 30-minute runtime limit or available memory budget are reported as \textit{N/A}. All experiments were performed on a Mac Mini M4 with 36 GB of unified memory.

    In the next subsection, we present our performance on publicly available data instances.

\subsection{CVRPLIB: AGS Benchmark for Large-Scale Instances}
    The AGS benchmark set contains some of the largest publicly available CVRP instances, with the sizes reaching approximately 30,000 customers, providing an important intermediate scale between traditional benchmarks and the industrial-scale CVRPs considered in this study \cite{arnold2019efficiently}.

    To evaluate decomposition performance beyond the original AGS benchmark while preserving realistic spatial and demand properties, we construct a collection of benchmark-derived instances by concatenating multiple AGS instances. For a batch size $b$, the generator combines the $b$ largest instances into a single instance. Two configurations are considered: \emph{Grid}, which preserves the spatial separation of the original instances to create large multi-region samples, and \emph{Unified Depot}, which merges all components into a single case sharing a common depot. This procedure substantially increases the scale while retaining the structural characteristics of the underlying benchmark data.

    When interpreting the results, miss rate should be viewed as the primary performance metric. In practical applications, failing to serve customer demand is typically far more consequential than increases in travel distance. Moreover, cost and miss rate are not independent objectives: omitting customers generally reduces the number of required visits and therefore lowers the total travel distance. Hence, methods exhibiting substantially higher miss rates may appear to achieve favorable routing costs even though the resulting solutions would require additional resources or delivery effort to reach comparable service levels.

    Tables~\ref{tab:ags_concat_2}, \ref{tab:ags_concat_6}, and \ref{tab:ags_concat_10} summarize the results on benchmark-derived AGS instances constructed using concatenation batch sizes of 2, 6, and 10, respectively.

    \begin{table}[h]
        \caption{Results on 2-instance concatenated AGS benchmark-derived instances. Created from Flanders1, Flanders2 with total 50,000 customers and 250 vehicles.}
        \label{tab:ags_concat_2}
        \centering
        
        \begin{minipage}{0.48\textwidth}
        \centering
        \subfloat[Grid + Original]{
        \scriptsize
        \begin{tabular}{|c|c|c|c|c|}
        \hline
        \rowcolor{gray!15}
        \textbf{Method} &
        \textbf{Time (s)} &
        \textbf{First Sol. (s)} &
        \textbf{Missed (\%)} &
        \textbf{Final Dist.} \\
        \hline
        BKM     & 0.275  & 35.191  & 3.980  & 722,038   \\ \hline
        CCBC    & 53.672 & 183.296 & 38.430 & 454,754   \\ \hline
        CC-CVRP & 6.137  & 35.539  & 4.230  & 685,687   \\ \hline
        ENT     & 0.020  & 34.967  & 4.270  & 716,555   \\ \hline
        GRD     & 0.030  & 35.601  & \textbf{0.000}  & 1,922,235 \\ \hline
        SWP     & 0.018  & 35.097  & 4.530  & 714,253   \\ \hline
        GLOP & -- & -- & \textbf{0.000} & 721,183 \\ \hline
        A-DEC   & 116.779 & 32.192 & \textbf{0.000} & 1,601,788 \\ \hline
        \end{tabular}
        }
        \end{minipage}

        \vspace{0.5em}
        \begin{minipage}{0.48\textwidth}
        \centering
        \subfloat[Grid + Scaled]{
        \scriptsize
        \begin{tabular}{|c|c|c|c|c|}
        \hline
        \rowcolor{gray!15}
        \textbf{Method} &
        \textbf{Time (s)} &
        \textbf{First Sol. (s)} &
        \textbf{Missed (\%)} &
        \textbf{Final Dist.} \\
        \hline
        BKM     & 0.274  & 35.153  & 3.980  & 728,020   \\ \hline
        CCBC    & 52.830 & 187.753 & 38.520 & 443,935   \\ \hline
        CC-CVRP & 6.305  & 35.316  & 4.140  & 706,542   \\ \hline
        ENT     & 0.020  & 35.075  & 4.060  & 715,728   \\ \hline
        GRD     & 0.031  & 35.213  & \textbf{0.020}  & 1,902,855 \\ \hline
        SWP     & 0.018  & 35.372  & 4.310  & 720,491   \\ \hline
        GLOP & -- & -- & \textbf{0.000} & 708,035 \\ \hline
        A-DEC   & 316.000 & 32.240  & 11.190 & 726,016  \\ \hline   
        \end{tabular}
        }
        \end{minipage}
        
        \vspace{0.5em}
        
        \begin{minipage}{0.48\textwidth}
        \centering
        \subfloat[Grid + Unified Depot]{
        \scriptsize
        \begin{tabular}{|c|c|c|c|c|}
        \hline
        \rowcolor{gray!15}
        \textbf{Method} &
        \textbf{Time (s)} &
        \textbf{First Sol. (s)} &
        \textbf{Missed (\%)} &
        \textbf{Final Dist.} \\
        \hline
        BKM     & 0.306  & 62.785  & 0.370  & 1,113,979 \\ \hline
        CCBC    & 42.164 & 193.401 & 36.190 & 445,526   \\ \hline
        CC-CVRP & 6.448  & 35.601  & 0.350  & 879,617   \\ \hline
        ENT     & 0.020  & 35.028  & 2.110  & 456,264   \\ \hline
        GRD     & 0.030  & 35.302  & 0.370  & 1,117,998 \\ \hline
        SWP     & 0.019  & 35.403  & 1.110  & 589,032   \\ \hline
        GLOP    & -- & -- & \textbf{0.000} & 631,661 \\ \hline
        A-DEC   & 319.175 & 32.209 & \textbf{0.000} & 860,906  \\ \hline   
        \end{tabular}
        }
        \end{minipage}
        
        \vspace{0.5em}
        
        \begin{minipage}{0.48\textwidth}
        \centering
        \subfloat[Overlapping]{
        \scriptsize
        \begin{tabular}{|c|c|c|c|c|}
        \hline
        \rowcolor{gray!15}
        \textbf{Method} &
        \textbf{Time (s)} &
        \textbf{First Sol. (s)} &
        \textbf{Missed (\%)} &
        \textbf{Final Dist.} \\
        \hline
        BKM     & 0.305  & 35.450  & 0.420  & 432,496   \\ \hline
        CCBC    & 58.705 & 172.150 & 38.790 & 192,314   \\ \hline
        CC-CVRP & 6.099  & 35.407  & 0.940  & 374,749   \\ \hline
        ENT     & 0.020  & 35.165  & 0.350  & 385,972   \\ \hline
        GRD     & 0.030  & 35.540  & \textbf{0.000}  & 1,027,172 \\ \hline
        SWP     & 0.018  & 35.547  & 0.560  & 411,024   \\ \hline
        GLOP & -- & -- & 2.250 & 2,395,735 \\ \hline
        A-DEC   & 316.004 & 32.235 & \textbf{0.000}  & 494,122   \\ \hline   
        \end{tabular}
        }
        \end{minipage}
    \end{table}

    \begin{table}[h]
        \caption{Results on 6-instance concatenated AGS benchmark-derived instances. Created from Flanders2, Flanders1, Brussels2, Brussels1, Ghent2, Ghent1, with total 102,000 customers and 655 vehicles.}
        \label{tab:ags_concat_6}
        \centering
        
        \begin{minipage}{0.48\textwidth}
        \centering
        \subfloat[Grid + Original]{
        \scriptsize
        \begin{tabular}{|c|c|c|c|c|}
        \hline
        \rowcolor{gray!15}
        \textbf{Method} &
        \textbf{Time (s)} &
        \textbf{First Sol. (s)} &
        \textbf{Missed (\%)} &
        \textbf{Final Dist.} \\
        \hline
        BKM     & 1.824   & 305.839 & 0.370 & 26,721,457  \\ \hline
        CCBC    & 153.129 & 1,068.166 & 28.400 & 6,952,275   \\ \hline
        CC-CVRP & 49.815  & 48.092  & 0.370 & 23,834,181  \\ \hline
        ENT     & 0.096   & 46.028  & 0.350 & 30,853,583  \\ \hline
        GRD     & 0.168   & 42.383  & 0.060 & 134,008,272 \\ \hline
        SWP     & 0.083   & 45.880  & 0.290 & 32,994,513  \\ \hline
        GLOP & -- & -- & N/A & N/A \\ \hline
        A-DEC   & 186.483 & 45.905  & \textbf{0.000}  & 16,856,072  \\ \hline   
        \end{tabular}
        }
        \end{minipage}
        
        \vspace{0.5em}
        
        \begin{minipage}{0.48\textwidth}
        \centering
        \subfloat[Grid + Scaled]{
        \scriptsize
        \begin{tabular}{|c|c|c|c|c|}
        \hline
        \rowcolor{gray!15}
        \textbf{Method} &
        \textbf{Time (s)} &
        \textbf{First Sol. (s)} &
        \textbf{Missed (\%)} &
        \textbf{Final Dist.} \\
        \hline
        BKM     & 1.839   & 306.428 & 0.280 & 27,224,341  \\ \hline
        CCBC    & 153.172 & 1,061.107 & 28.430 & 6,772,262   \\ \hline
        CC-CVRP & 51.114  & 45.616  & 0.300 & 24,554,350  \\ \hline
        ENT     & 0.094   & 48.711  & 0.260 & 29,982,591  \\ \hline
        GRD     & 0.169   & 43.706  & \textbf{0.060} & 133,716,715 \\ \hline
        SWP     & 0.084   & 45.905  & 0.230 & 33,715,569  \\ \hline
        GLOP & -- & -- & N/A & N/A \\ \hline
        A-DEC   & 207.984 & 45.776  & \textbf{0.040} & 16,572,978  \\ \hline   
        \end{tabular}
        }
        \end{minipage}
        
        \vspace{0.5em}
        
        \begin{minipage}{0.48\textwidth}
        \centering
        \subfloat[Grid + Unified Depot]{
        \scriptsize
        \begin{tabular}{|c|c|c|c|c|}
        \hline
        \rowcolor{gray!15}
        \textbf{Method} &
        \textbf{Time (s)} &
        \textbf{First Sol. (s)} &
        \textbf{Missed (\%)} &
        \textbf{Final Dist.} \\
        \hline
        BKM     & 8.755   & 303.392 & 0.000 & 31,290,192  \\ \hline
        CCBC    & 151.896 & 1,093.719 & 14.480 & 6,999,298   \\ \hline
        CC-CVRP & 45.569  & 43.194  & \textbf{0.010} & 24,099,312  \\ \hline
        ENT     & 0.091   & 42.753  & 0.270 & 23,725,871  \\ \hline
        GRD     & 0.866   & 91.552  & \textbf{0.000} & 94,869,682  \\ \hline
        SWP     & 0.083   & 43.475  & 0.080 & 30,000,951  \\ \hline
        GLOP & -- & -- & N/A & N/A \\ \hline
        A-DEC   & 187.599 & 44.055  & 0.480 & 15,681,319  \\ \hline   
        \end{tabular}
        }
        \end{minipage}
        
        \vspace{0.5em}
        
        \begin{minipage}{0.48\textwidth}
        \centering
        \subfloat[Overlapping]{
        \scriptsize
        \begin{tabular}{|c|c|c|c|c|}
        \hline
        \rowcolor{gray!15}
        \textbf{Method} &
        \textbf{Time (s)} &
        \textbf{First Sol. (s)} &
        \textbf{Missed (\%)} &
        \textbf{Final Dist.} \\
        \hline
        BKM     & 15.100  & 120.460 & 0.100 & 11,314,913  \\ \hline
        CCBC    & 150.762 & 1,046.884 & 28.180 & 1,668,047   \\ \hline
        CC-CVRP & 43.539  & 45.575  & 0.080 & 9,750,538   \\ \hline
        ENT     & 0.092   & 43.938  & 0.020 & 10,867,047  \\ \hline
        GRD     & 0.876   & 112.935 & 0.050 & 34,320,349  \\ \hline
        SWP     & 0.085   & 46.338  & 0.030 & 15,681,932  \\ \hline
        GLOP & -- & -- & N/A & N/A \\ \hline
        A-DEC   & 226.630 & 44.861  & \textbf{0.000} & 8,284,592   \\ \hline   
        \end{tabular}
        }
        \end{minipage}
    \end{table}

    \begin{table}[h]
        \caption{Results on 10-instance concatenated AGS benchmark-derived instances. Created from instances Flanders2, Flanders1, Brussels2, Brussels1, Ghent2, Ghent1, Antwerp2, Antwerp1, Leuven2, Leuven1 with total customers 122,000 and vehicles 960.}
        \label{tab:ags_concat_10}
        \centering
        
        \begin{minipage}{0.48\textwidth}
        \centering
        \subfloat[Grid + Original]{
        \scriptsize
        \begin{tabular}{|c|c|c|c|c|}
        \hline
        \rowcolor{gray!15}
        \textbf{Method} &
        \textbf{Time (s)} &
        \textbf{First Sol. (s)} &
        \textbf{Missed (\%)} &
        \textbf{Final Dist.} \\
        \hline
        BKM     & 16.900  & 346.626  & 3.220  & 323,815,921   \\ \hline
        CCBC    & 164.035 & 2199.457 & 31.790 & 241,436,895   \\ \hline
        CC-CVRP & 114.813 & 56.844   & 3.210  & 313,229,684   \\ \hline
        ENT     & 0.181   & 54.748   & 3.220  & 362,590,300   \\ \hline
        GRD     & 3.947   & 169.017  & 0.020  & 4,883,751,464 \\ \hline
        SWP     & 0.163   & 55.966   & 3.190  & 958,394,465   \\ \hline
        GLOP & -- & -- & N/A & N/A \\ \hline
        A-DEC   & 154.321 & 87.369   & \textbf{0.000}  & 356,963,206 \\ \hline   
        \end{tabular}
        }
        \end{minipage}
        
        \vspace{0.5em}
        
        \begin{minipage}{0.48\textwidth}
        \centering
        \subfloat[Grid + Scaled]{
        \scriptsize
        \begin{tabular}{|c|c|c|c|c|}
        \hline
        \rowcolor{gray!15}
        \textbf{Method} &
        \textbf{Time (s)} &
        \textbf{First Sol. (s)} &
        \textbf{Missed (\%)} &
        \textbf{Final Dist.} \\
        \hline
        BKM     & 4.732   & 303.940  & 2.380  & 549,358,450   \\ \hline
        CCBC    & 152.548 & 2276.662 & 31.630 & 218,822,415   \\ \hline
        CC-CVRP & 125.363 & 80.223   & 2.540  & 564,676,075   \\ \hline
        ENT     & 0.196   & 80.353   & 2.210  & 589,525,949   \\ \hline
        GRD     & 0.379   & 58.736   & 0.020  & 4,369,090,450 \\ \hline
        SWP     & 0.172   & 58.594   & 2.310  & 1,283,021,462 \\ \hline
        GLOP & -- & -- & N/A & N/A \\ \hline
        A-DEC   & 219.875 & 65.293   & \textbf{0.000}  & 525,301,597 \\ \hline   
        \end{tabular}
        }
        \end{minipage}
        
        \vspace{0.5em}
        
        \begin{minipage}{0.48\textwidth}
        \centering
        \subfloat[Grid + Unified Depot]{
        \scriptsize
        \begin{tabular}{|c|c|c|c|c|}
        \hline
        \rowcolor{gray!15}
        \textbf{Method} &
        \textbf{Time (s)} &
        \textbf{First Sol. (s)} &
        \textbf{Missed (\%)} &
        \textbf{Final Dist.} \\
        \hline
        BKM     & 4.838   & 82.656   & 1.640  & 781,696,307   \\ \hline
        CCBC    & 152.961 & 2138.380 & 7.640  & 225,650,614   \\ \hline
        CC-CVRP & 113.079 & 57.250   & 1.750  & 569,001,356   \\ \hline
        ENT     & 0.189   & 55.703   & 1.110  & 574,290,940   \\ \hline
        GRD     & 0.382   & 56.634   & \textbf{0.000}  & 3,531,830,252 \\ \hline
        SWP     & 0.168   & 58.153   & 1.310  & 888,401,316   \\ \hline
        GLOP & -- & -- & N/A & N/A \\ \hline
        A-DEC   & 217.488 & 72.279   & \textbf{0.000}  & 387,027,477 \\ \hline   
        \end{tabular}
        }
        \end{minipage}
        
        \vspace{0.5em}
        
        \begin{minipage}{0.48\textwidth}
        \centering
        \subfloat[Overlapping]{
        \scriptsize
        \begin{tabular}{|c|c|c|c|c|}
        \hline
        \rowcolor{gray!15}
        \textbf{Method} &
        \textbf{Time (s)} &
        \textbf{First Sol. (s)} &
        \textbf{Missed (\%)} &
        \textbf{Final Dist.} \\
        \hline
        BKM     & 5.404   & 58.432   & 0.600  & 132,992,549   \\ \hline
        CCBC    & 155.899 & 2256.370 & 31.600 & 18,601,621    \\ \hline
        CC-CVRP & 116.655 & 55.987   & 0.690  & 113,854,206   \\ \hline
        ENT     & 0.182   & 60.752   & 0.550  & 134,914,301   \\ \hline
        GRD     & 0.386   & 57.370   & \textbf{0.000}  & 472,584,108   \\ \hline
        SWP     & 0.169   & 58.108   & 0.450  & 234,799,506   \\ \hline
        GLOP & -- & -- & N/A & N/A \\ \hline
        A-DEC   & 233.900 & 58.526   & \textbf{0.000}  & 83,125,139  \\ \hline   
        \end{tabular}
        }
        \end{minipage}
    \end{table}

    Among the optimization-based methods, CCBC achieves the lowest distances but at the cost of substantially higher miss rates, often exceeding 25\% of total demand. Since unserved customers directly reduce the amount of routing required, these distances are not directly comparable to methods achieving near-complete demand coverage. In contrast, BKM, ENT, and CC-CVRP maintain substantially lower miss rates, with CC-CVRP generally providing the strongest balance between service quality and cost within this class.

    The geometric methods exhibit mixed behavior. Grid-based clustering frequently delivers near-zero miss rates because the resulting clusters remain geographically coherent and easy to route independently. However, this comes at the expense of extremely large distances, particularly as the scale increases. Sweep-based clustering generally produces lower costs than GRD while maintaining comparable miss rates.
    
    The learning-based GLOP baseline attains favorable performance on the 50,000-customer instances but fails to produce valid solutions for the larger 102,000- and 122,000-customer benchmarks. This highlights the difficulty of scaling existing learning-based routing approaches to the size levels considered in this study.
    
    The proposed A-DEC model exhibits a different scaling trend. While its performance on the two-instance benchmarks is broadly comparable to existing methods, its advantage becomes increasingly apparent as the problem size grows. For the six-instance and ten-instance benchmarks, A-DEC achieves among the lowest distances while maintaining near-zero miss rates, often outperforming both geometric and optimization-based alternatives.

    In the next section, we present the robustness analysis we test our algorithm on various dimensions including customer count, vehicle count, different depot and demand patterns.
    
\subsection{Large-Scale Robustness Analysis}
    While the AGS benchmark set contains some of the largest standardized CVRP instances, reaching approximately 30,000 customers, the primary objective of this work is to evaluate decomposition performance at substantially larger scales. To this end, we generate synthetic CVRP instances containing up to several hundred thousand customers while systematically varying key configuration dimensions. Each instance name encodes the generation parameters used to create the corresponding dataset, enabling controlled ablation studies over individual dimensions.

    The generation parameters are summarized below:
    \begin{itemize}
        \item \textbf{$n$ (Customer Count):} Total number of customers.
        \item \textbf{$k$ (Vehicle Count):} Total number of available vehicles.
        \item \textbf{$D$ (Demand Pattern):} Distribution of customer demands.
        \item \textbf{$M$ (Depot Configuration):} Number and placement of depot locations.
    \end{itemize}

    In addition to the baseline decomposition methods, we evaluate two variants of our approach using the best-performing language models identified during preliminary experiments, denoted \textbf{Qwen2.5} \cite{qwen2025qwen25technicalreport} and \textbf{Qwen3} \cite{yang2025qwen3technicalreport}.

    In this section, we pick few best-performing models from each type of clustering paradigm to compare.

\subsubsection{Customer Count Variation}
\begin{figure}[h]
        \centering
        \subfloat[Clustering time]{
            \includegraphics[width=0.48\textwidth]{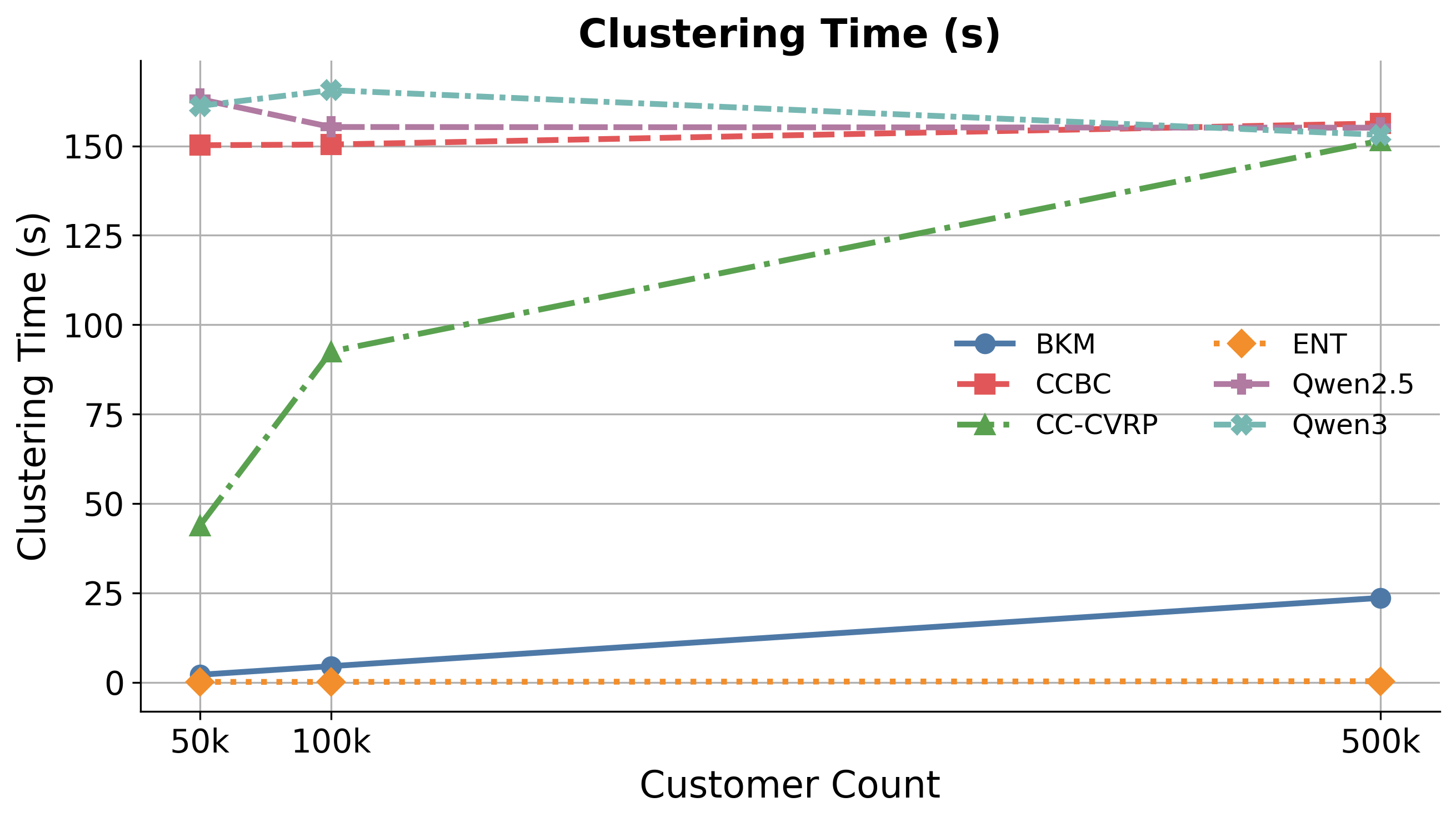}
        }\hfill
        \subfloat[Initial miss rate]{
            \includegraphics[width=0.48\textwidth]{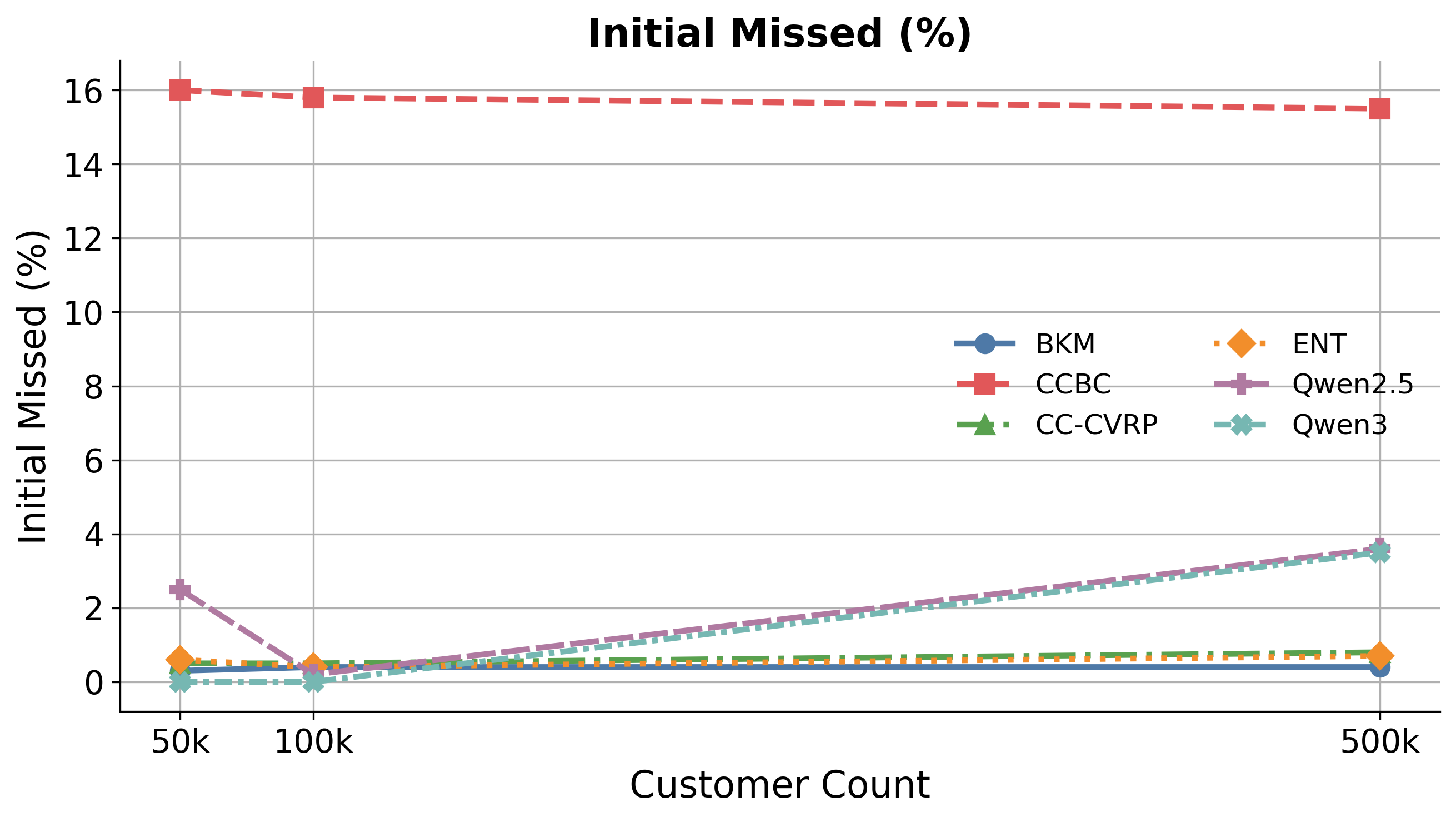}
        }
        \vspace{0.5em}
        \subfloat[Final miss rate]{
            \includegraphics[width=0.48\textwidth]{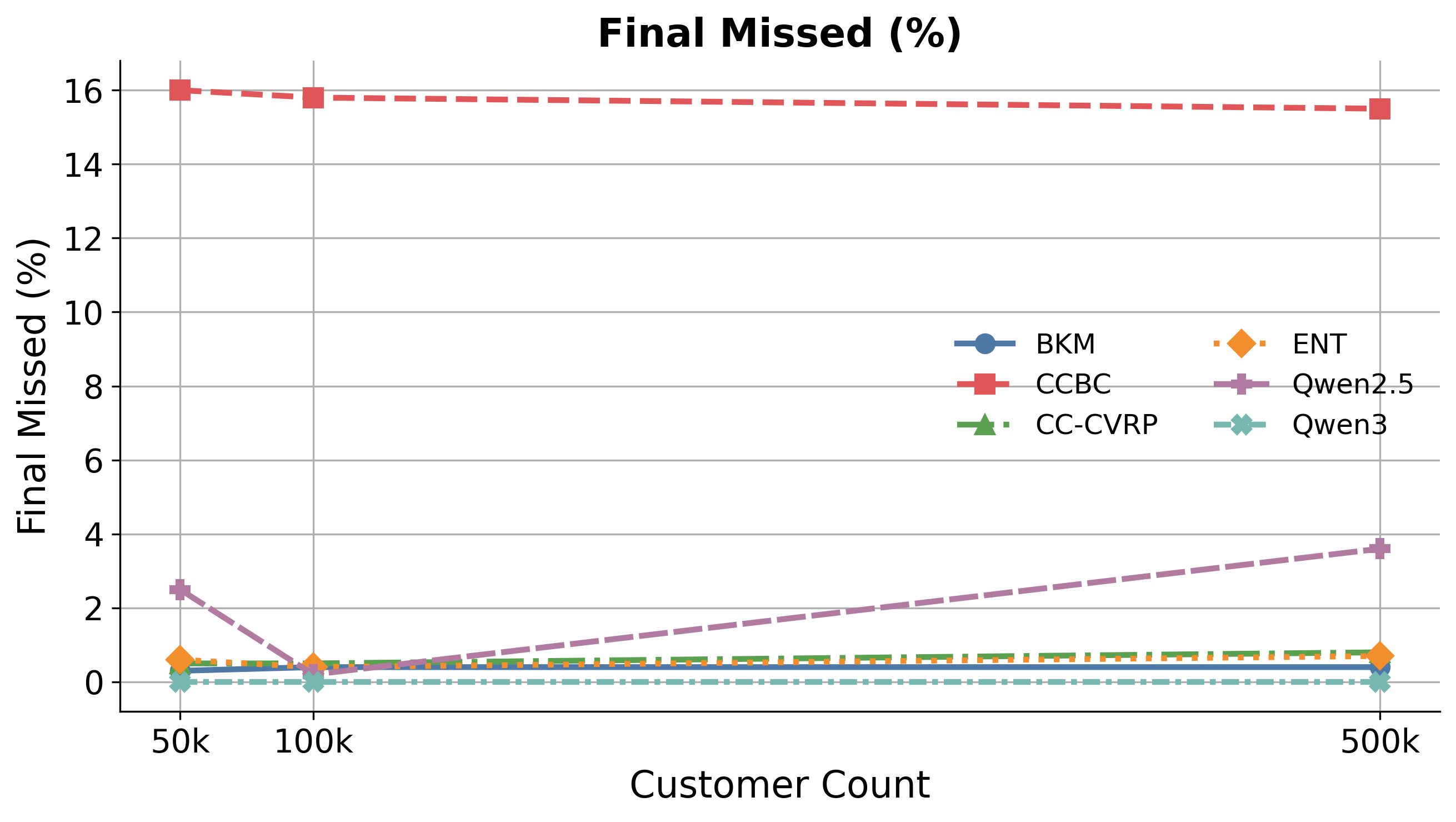}
        }\hfill
        \subfloat[Final routing distance]{
            \includegraphics[width=0.48\textwidth]{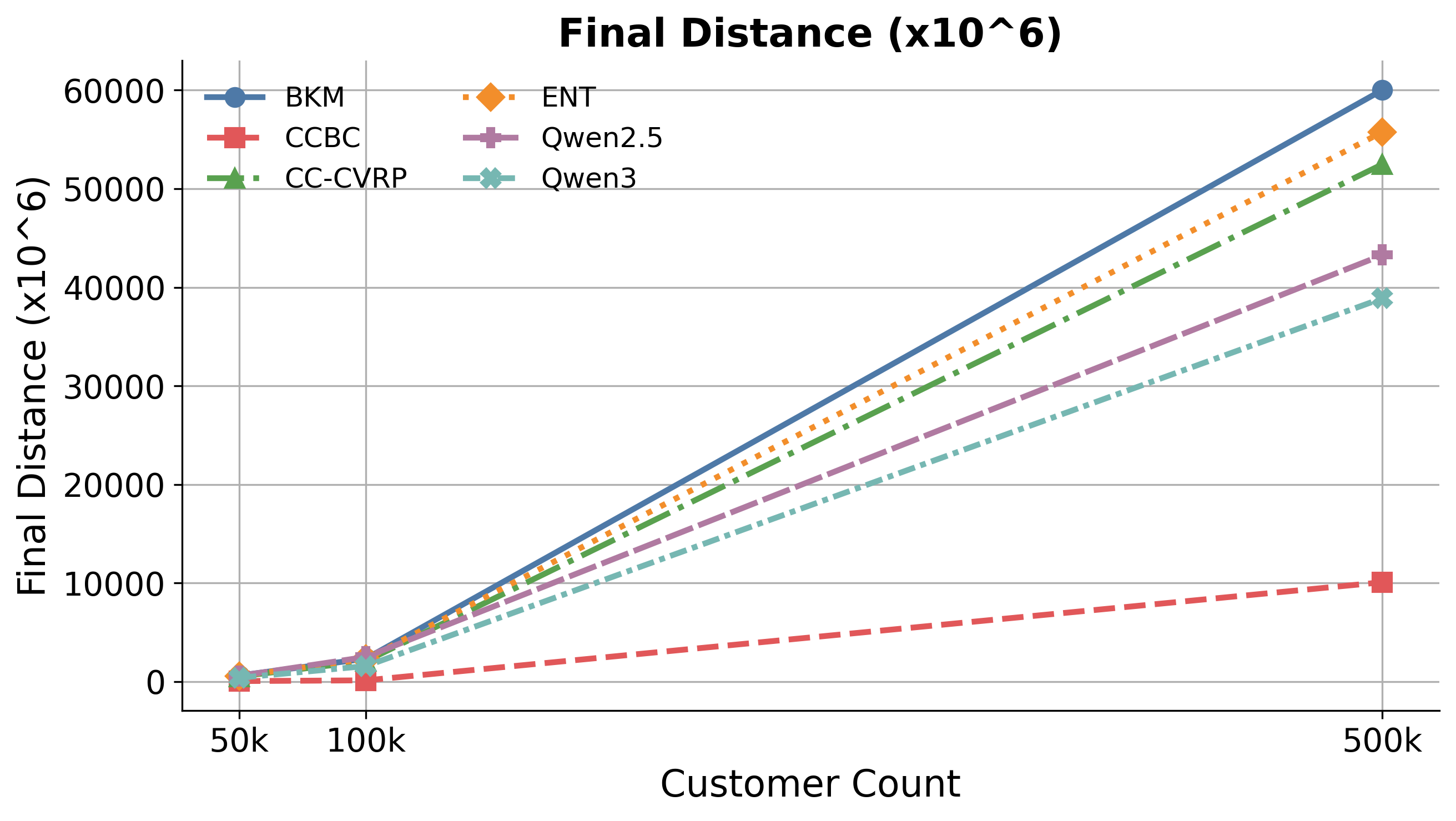}
        }
        \caption{Customer count variation results. Each plot shows the evolution of one performance metric as the number of customers increases while all remaining generation parameters are held fixed.}
        \label{fig:customer_count_variation}
    \end{figure}

    The customer count variation evaluates the scalability of methods as the number of customers increases while all remaining generation parameters are held fixed. Figure~\ref{fig:customer_count_variation} provides a visual comparison of the principal downstream metrics.

    Figure~\ref{fig:customer_count_variation} summarizes the effect of increasing customer count on decomposition performance. Overall, our methodology remains competitive as problem size increases, with the Qwen3 variant consistently achieving the lowest distance while maintaining a zero final miss rate throughout all evaluated customer counts. In contrast, CCBC attains substantially lower distances at the expense of high customer miss rates, while conventional geometric and optimization-based methods generally preserve low miss rates but incur considerably higher distances as the scale grows. These results reveal that the proposed adaptive decomposition strategy scales effectively from benchmark-scale instances to industrial-scale routing containing up to 500,000 customers.

\subsubsection{Vehicle Count Variation}
\begin{figure}[h]
        \centering
        \subfloat[Clustering time]{
            \includegraphics[width=0.48\textwidth]{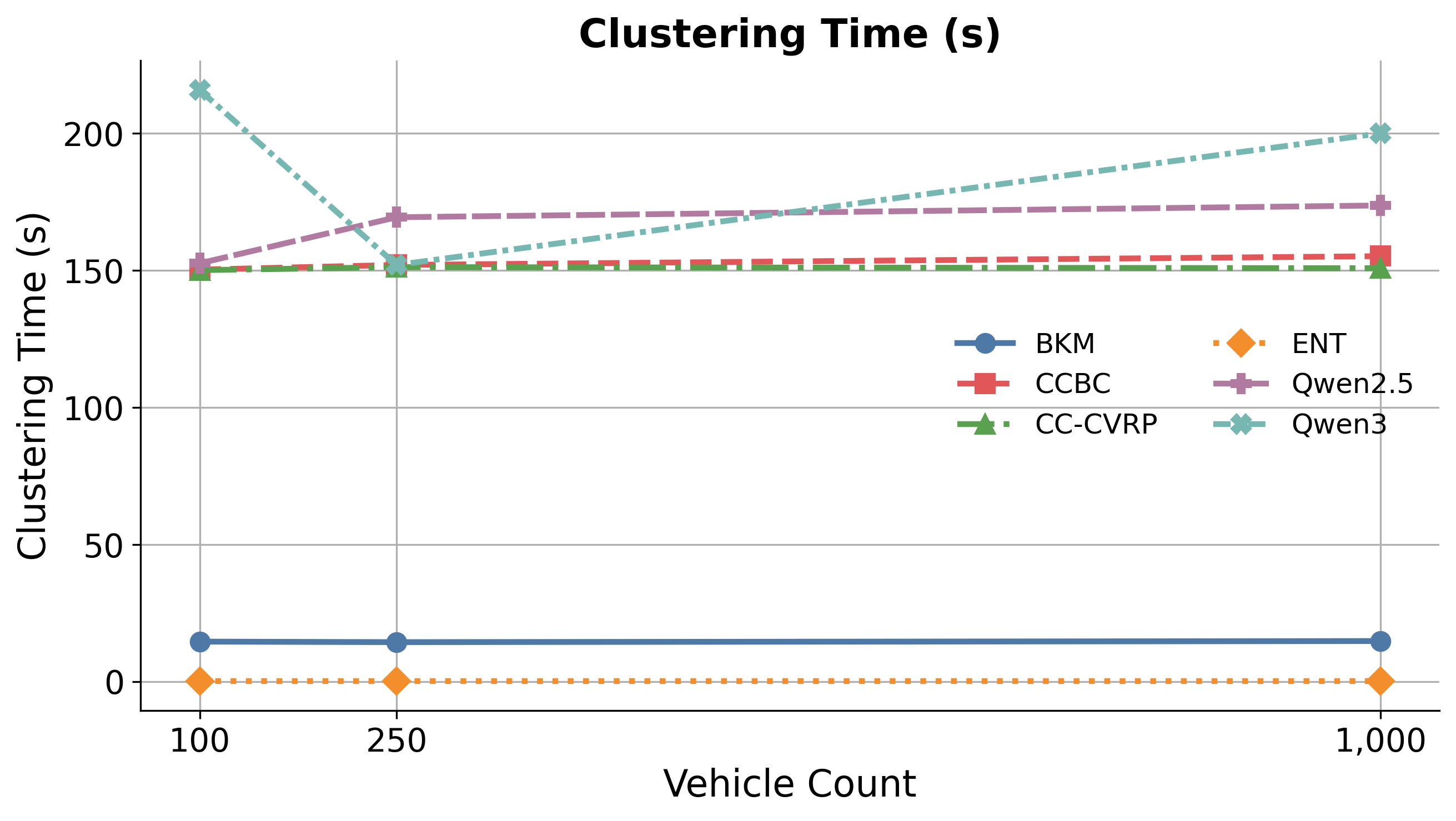}
        }\hfill
        \subfloat[Initial miss rate]{
            \includegraphics[width=0.48\textwidth]{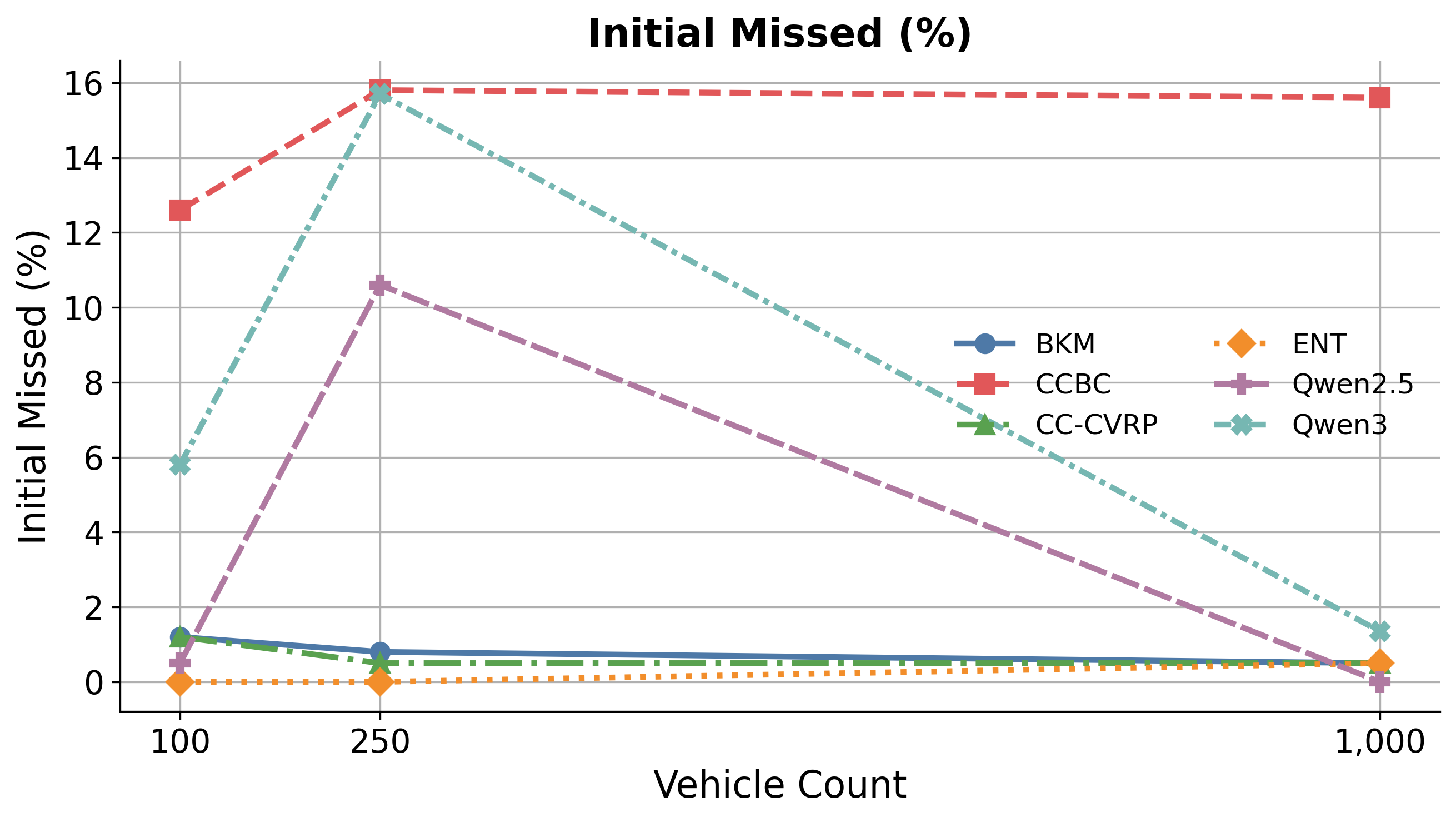}
        }
        \vspace{0.5em}
        \subfloat[Final miss rate]{
            \includegraphics[width=0.48\textwidth]{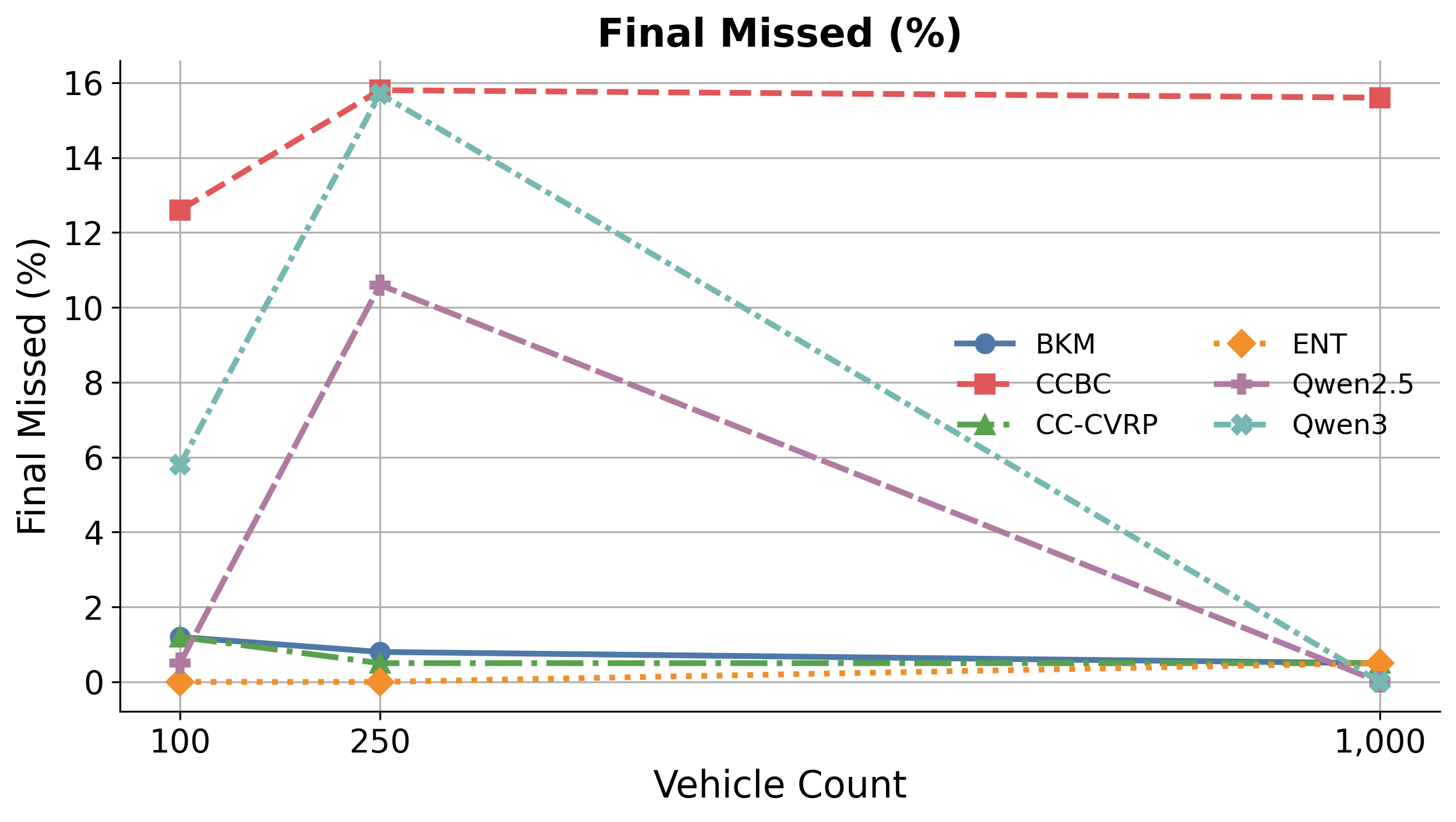}
        }\hfill
        \subfloat[Final routing distance]{
            \includegraphics[width=0.48\textwidth]{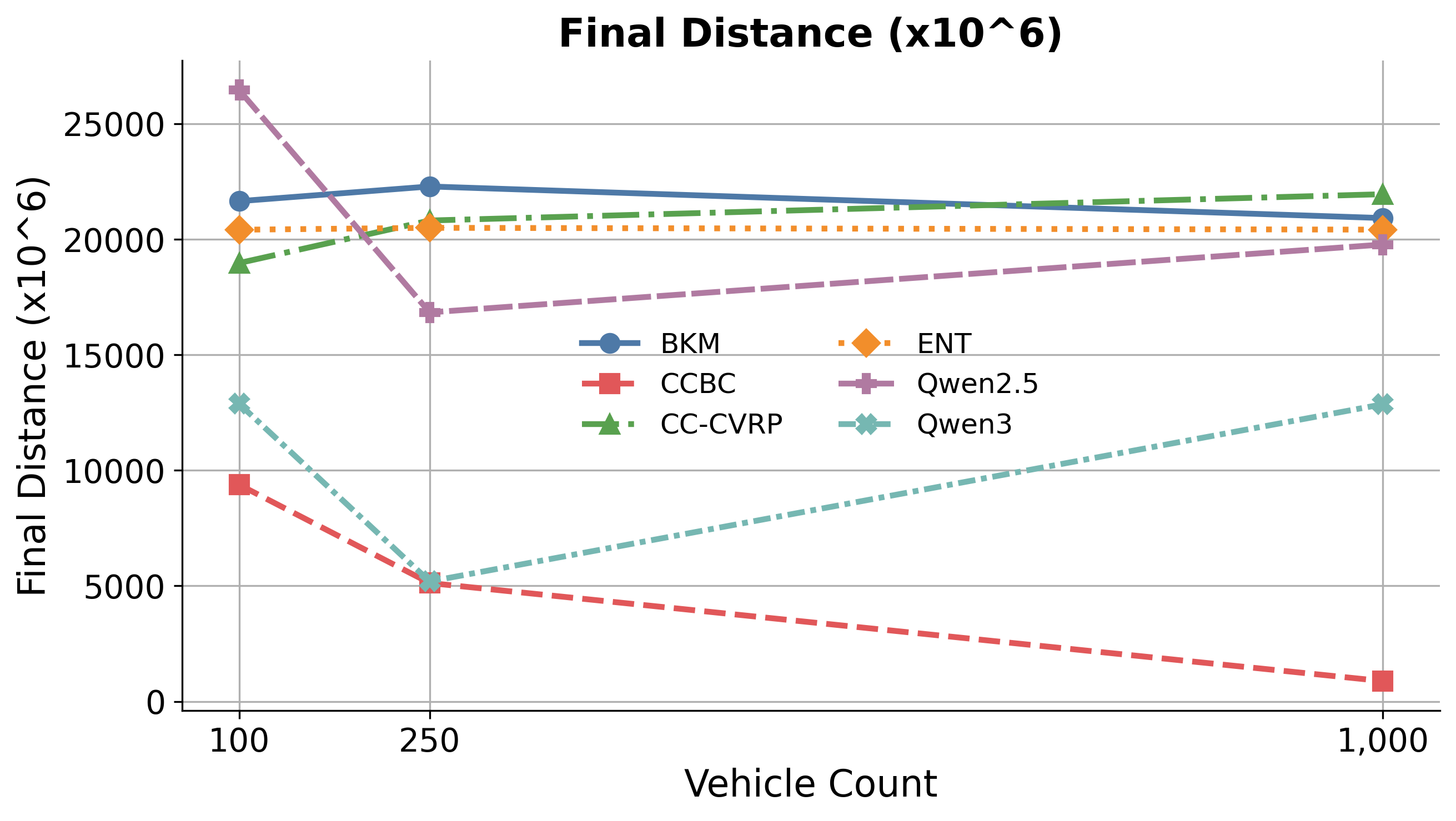}
        }
        \caption{Vehicle count variation results. Each plot illustrates the effect of varying the number of available vehicles on one downstream performance metric while keeping all remaining generation parameters fixed.}
        \label{fig:vehicle_count_variation}
    \end{figure}

    The vehicle count variation evaluates the sensitivity of methods to fleet size by varying the number of available vehicles while keeping all remaining generation parameters fixed. The resulting performance across the evaluated vehicle counts is reported in Figure~\ref{fig:vehicle_count_variation}.

    Figure~\ref{fig:vehicle_count_variation} exhibits trends consistent with the customer count analysis. Under all the evaluated fleet sizes, the proposed adaptive decomposition framework maintains competitive operational performance, with the Qwen3 variant achieving the lowest distance while maintaining low final miss rates. In particular, for 1,000 vehicles, Qwen3 fully recovers all initially missed demand and reduces distance by approximately 40\% relative to the strongest non-adaptive decomposition baselines. This suggests that the decomposition methodology adapts effectively to varying levels of available routing resources.

\subsubsection{Demand Pattern Variation}
\begin{figure*}[h]
        \centering
        \subfloat[Clustering time]{
            \includegraphics[width=0.48\textwidth]{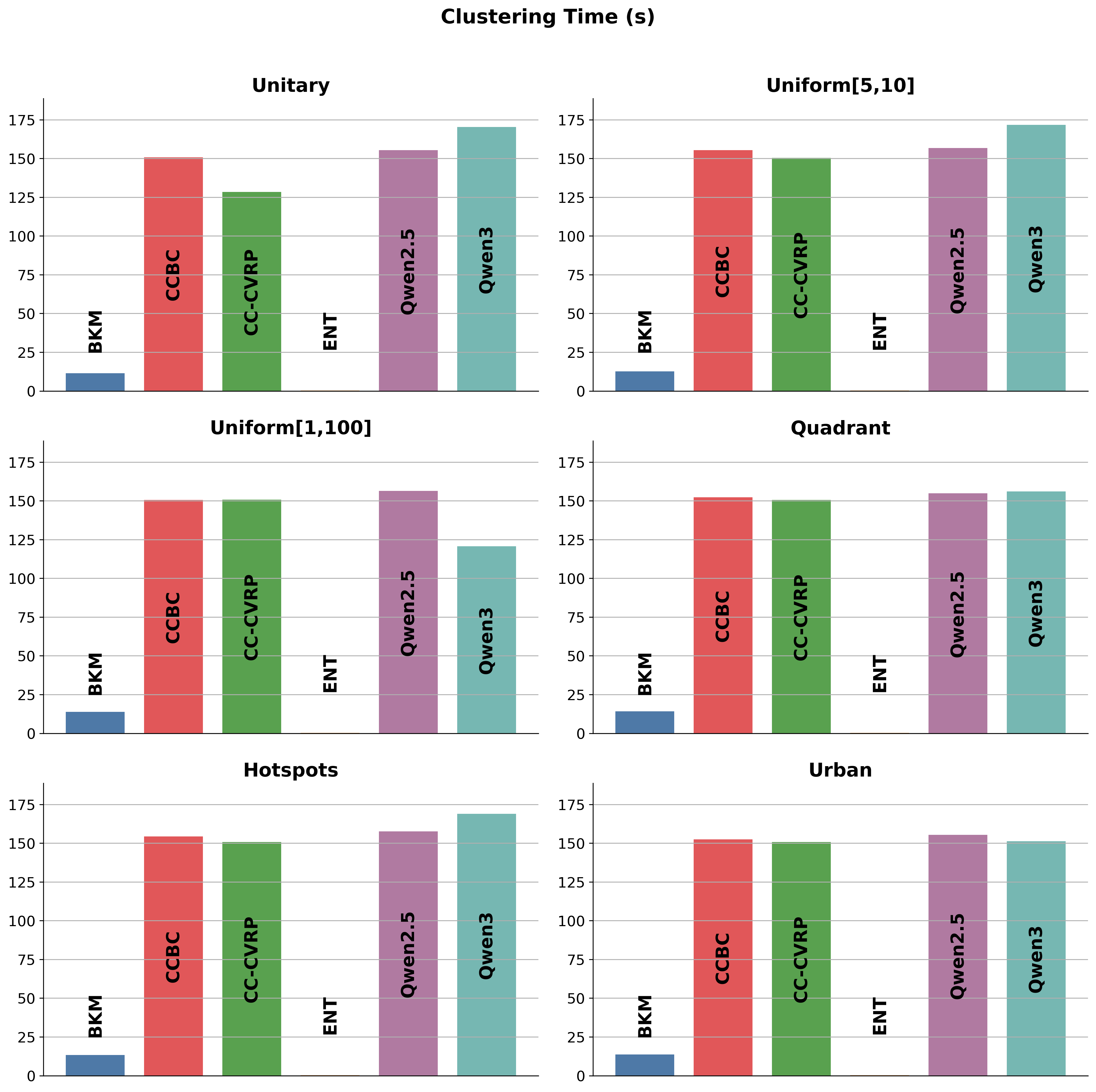}
        }\hfill
        \subfloat[Initial miss rate]{
            \includegraphics[width=0.48\textwidth]{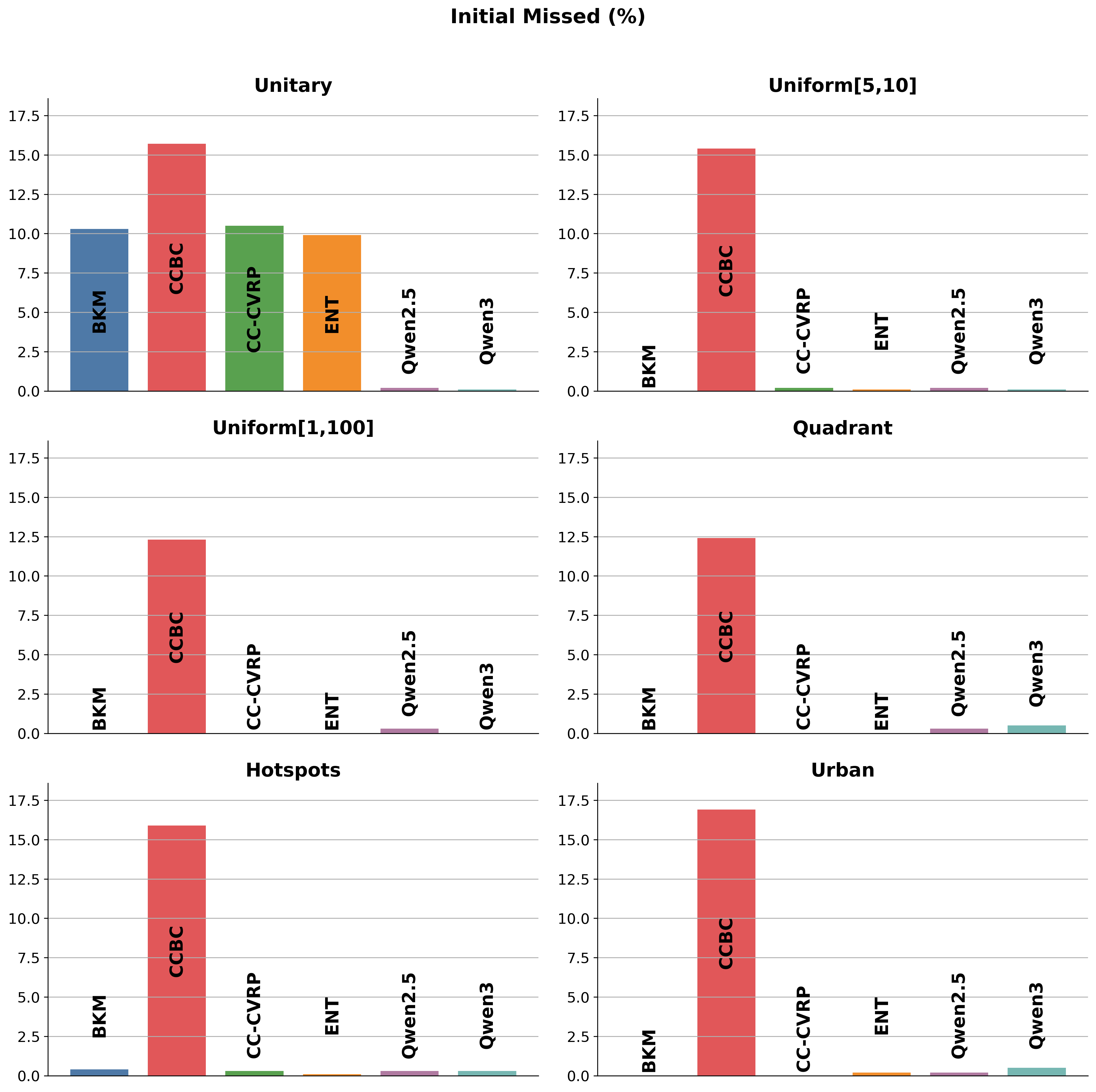}
        }
    
        \vspace{0.5em}
    
        \subfloat[Final miss rate]{
            \includegraphics[width=0.48\textwidth]{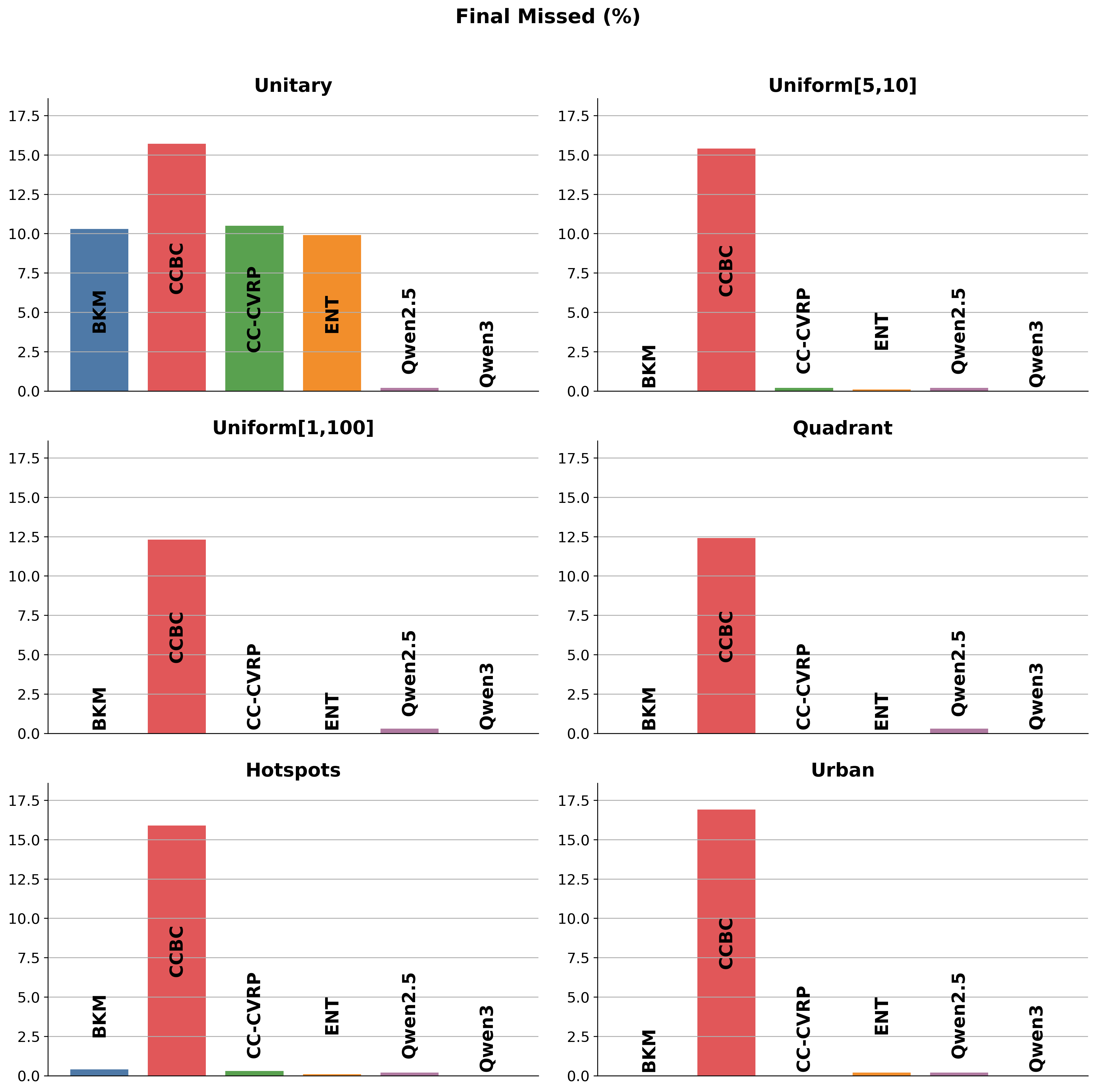}
        }\hfill
        \subfloat[Final routing distance]{
            \includegraphics[width=0.48\textwidth]{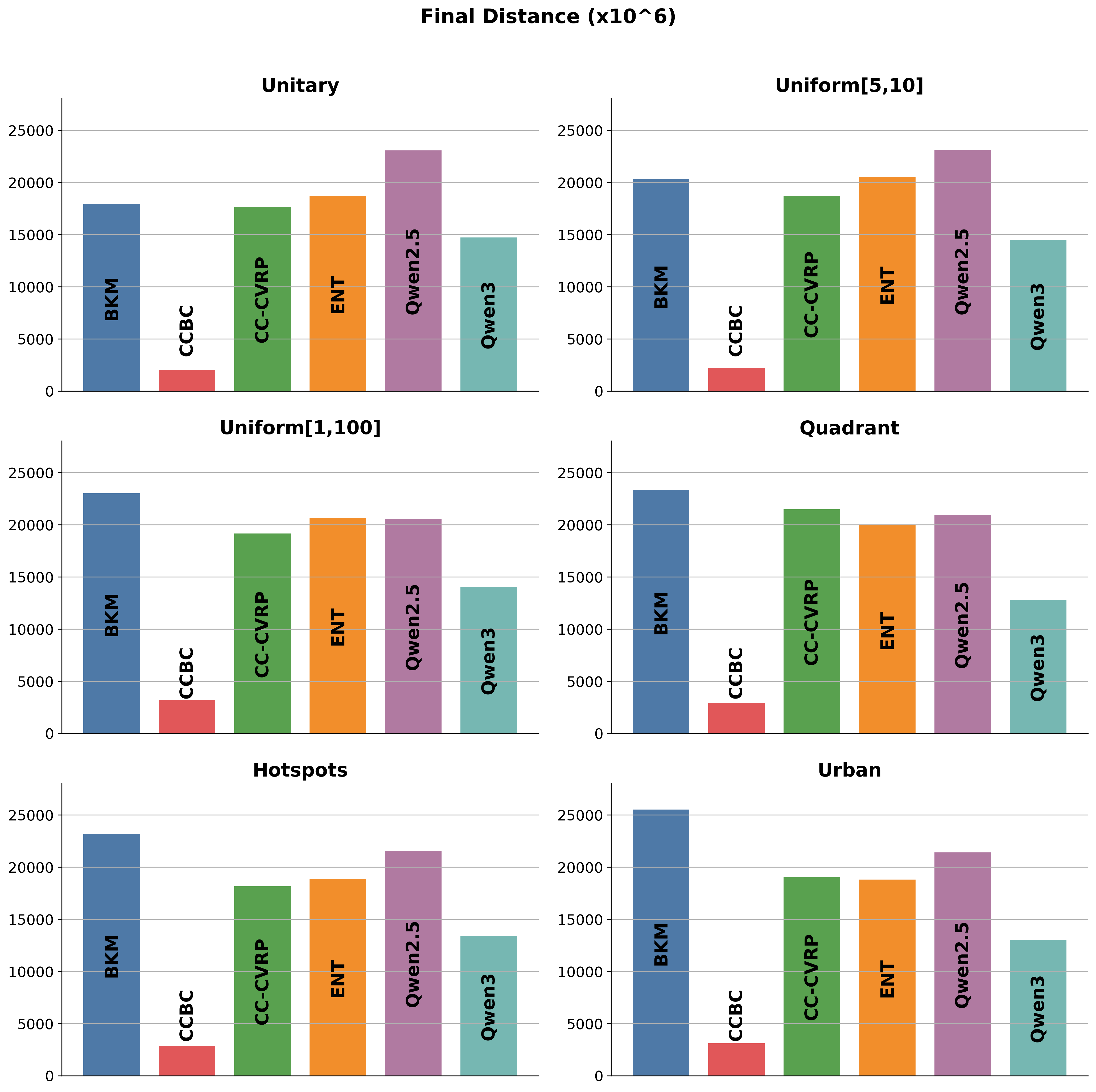}
        }
    
        \caption{Demand pattern variation results. Each plot compares the performance of the evaluated decomposition methods across six demand-generation patterns while keeping all remaining generation parameters fixed.}
        \label{fig:demand_pattern_variation}
    \end{figure*}

    In this subsection, we evaluate the demand pattern variation robustness of decomposition methods to different customer demand distributions. The resulting performance over the evaluated demand patterns is reported in Figure~\ref{fig:demand_pattern_variation}.

    The evaluated demand patterns introduce progressively more challenging demand structures. \textit{Unitary} and \textit{Uniform} distributions provide relatively homogeneous demand levels, while \textit{Quadrant} introduces significant spatial variation in demand intensity. \textit{Hotspots} and \textit{Urban} generate localized regions of elevated demand, creating demand-capacity imbalances that may be difficult to capture through purely geometric strategies.

    Figure~\ref{fig:demand_pattern_variation} illustrates that our model remains robust under diverse customer demand distributions. Similar to the customer and vehicle count analyses, the Qwen3 variant consistently producing the lowest routing distance while maintaining a zero final miss rate for all evaluated demand patterns. These results suggest that the proposed adaptive decomposition strategy effectively accommodates both uniform and highly heterogeneous demand patterns encountered in very large problems.

\subsubsection{Depot Configuration Variation}
\begin{figure}[h]
        \centering
        \subfloat[Clustering time]{
            \includegraphics[width=0.48\textwidth]{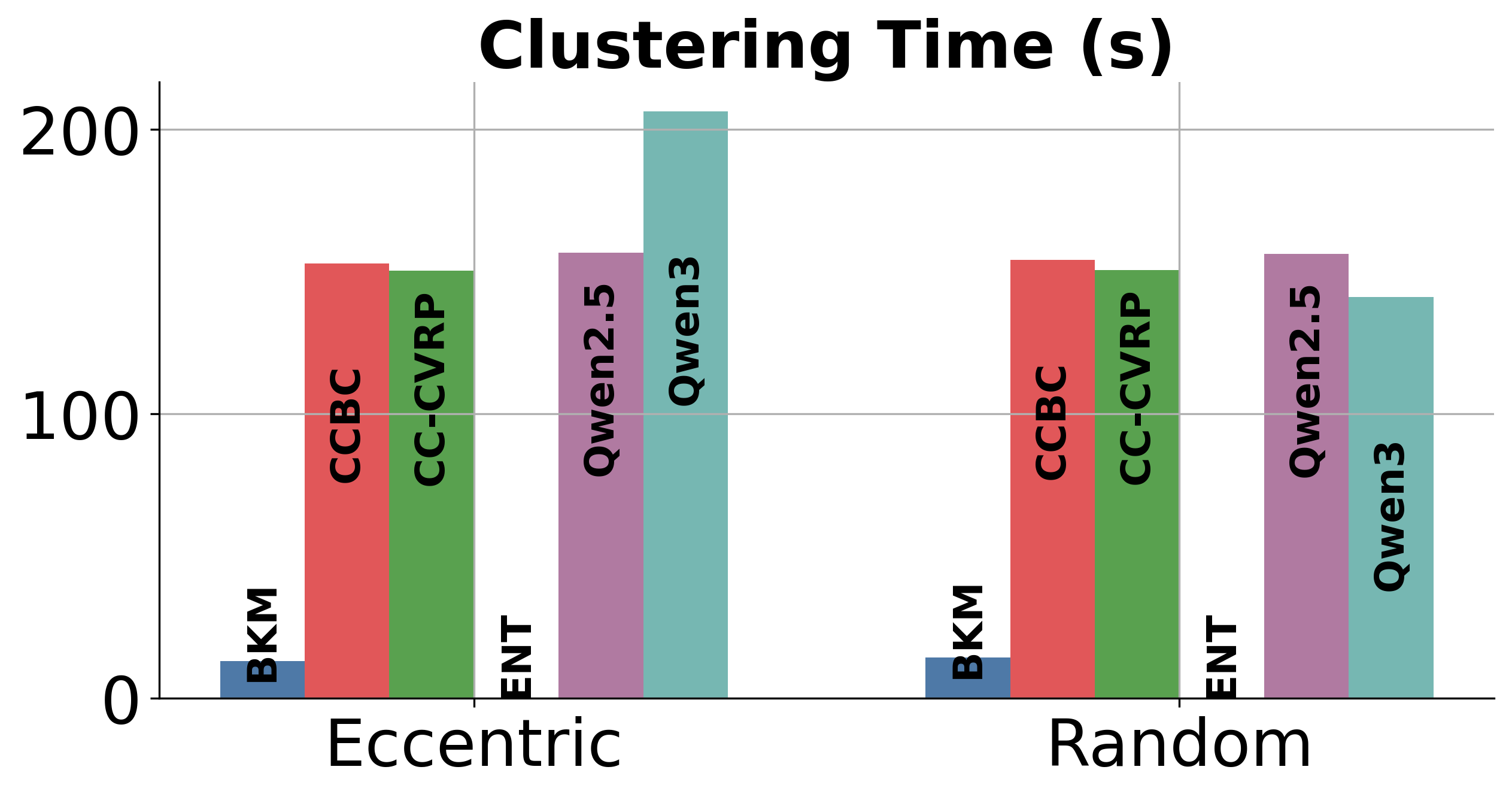}
        }\hfill
        \subfloat[Initial miss rate]{
            \includegraphics[width=0.48\textwidth]{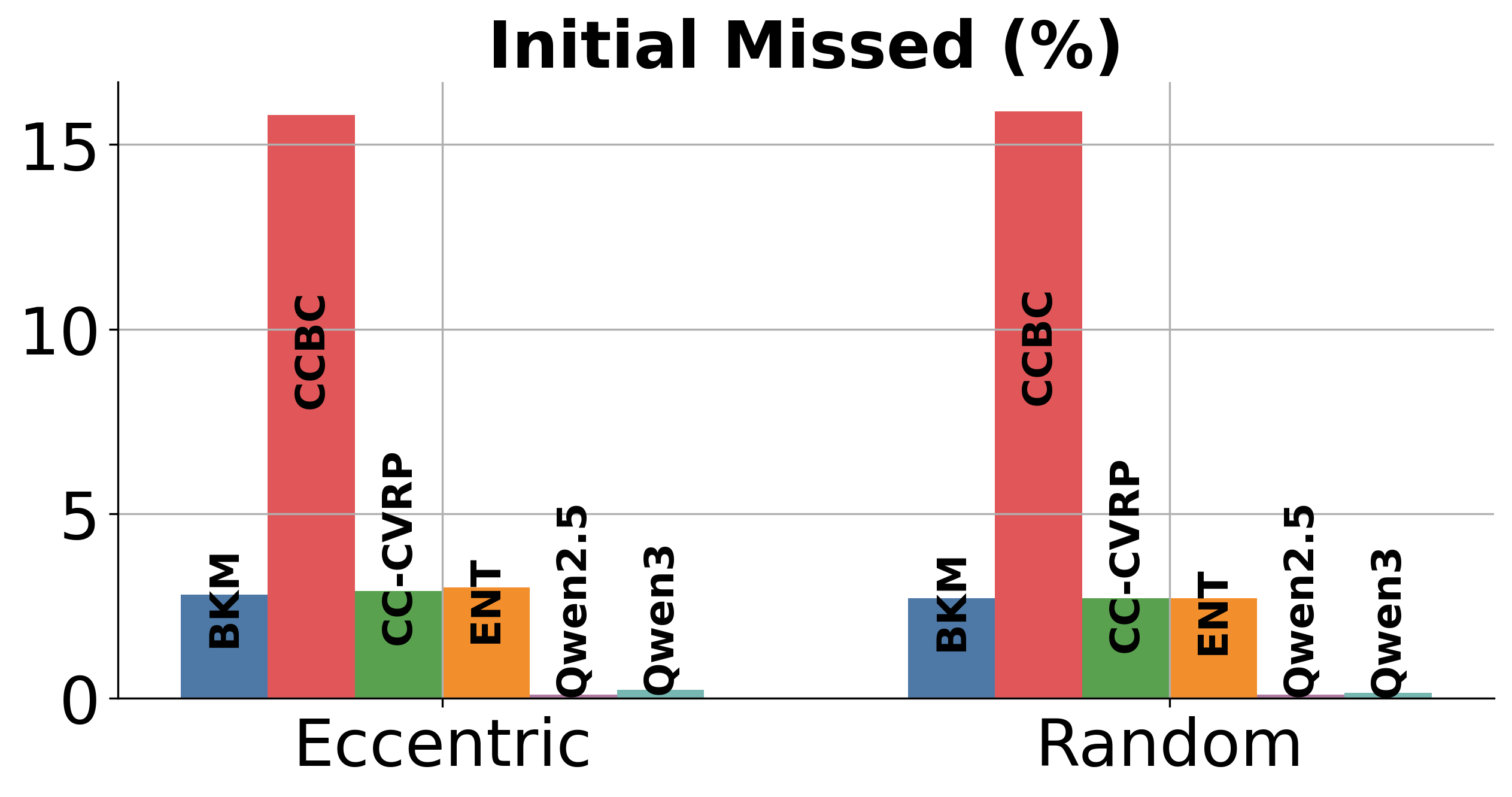}
        }
        \vspace{0.5em}
        \subfloat[Final miss rate]{
            \includegraphics[width=0.48\textwidth]{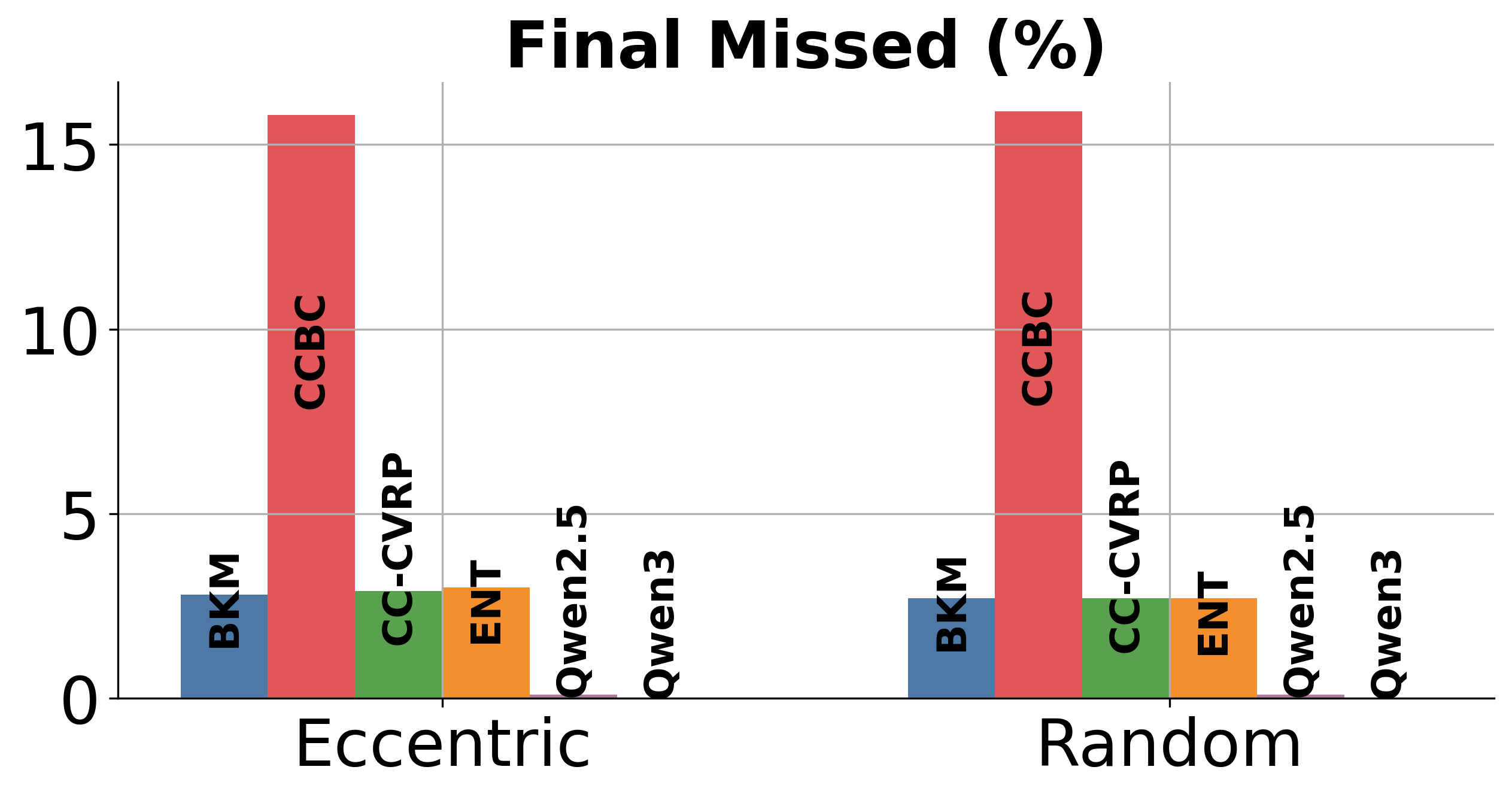}
        }\hfill
        \subfloat[Final routing distance]{
            \includegraphics[width=0.48\textwidth]{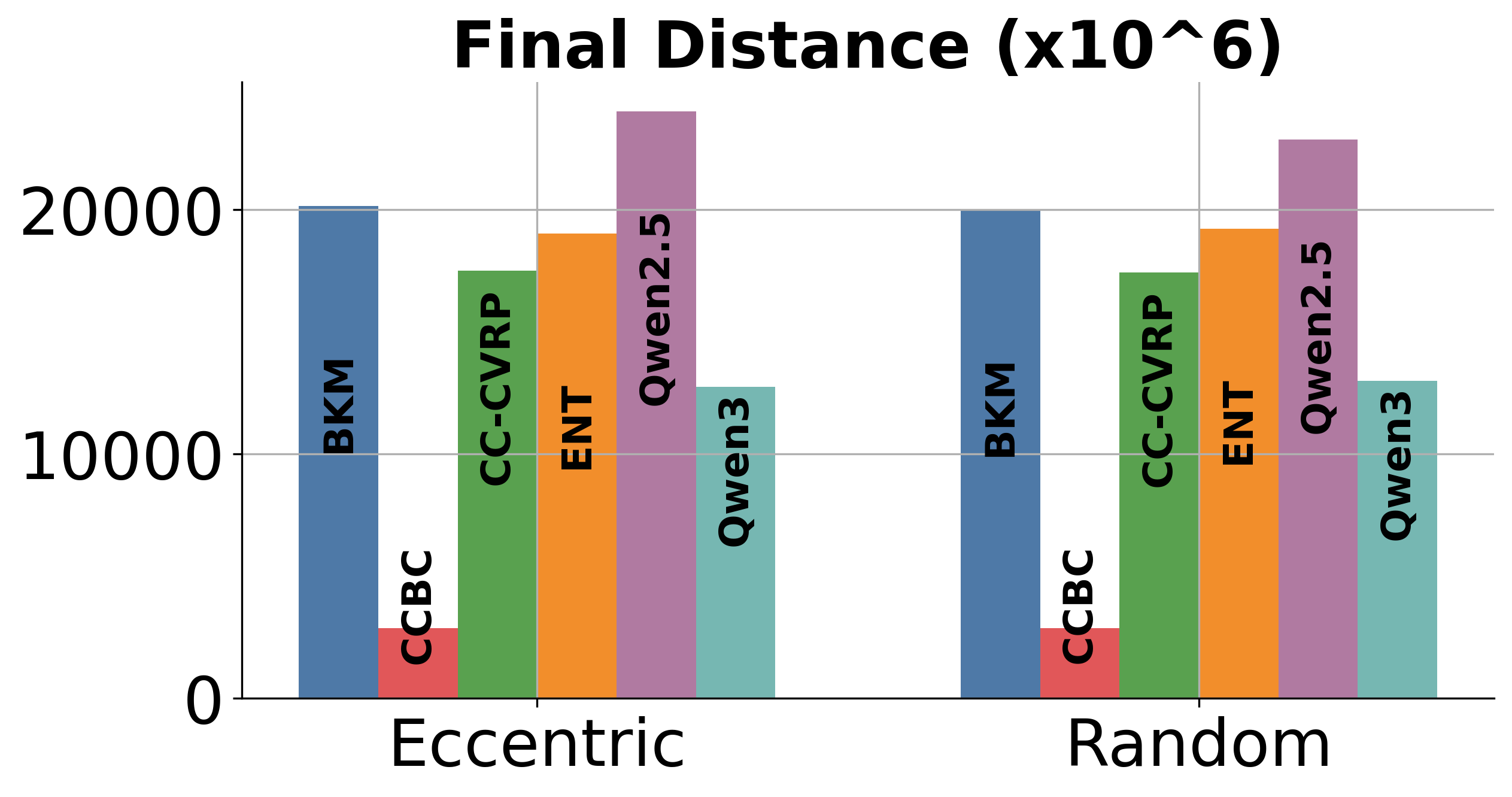}
        }
        \caption{Depot configuration variation results. Each plot compares the effect of eccentric and random depot placement on one downstream performance metric while keeping all remaining generation parameters fixed.}
    
        \label{fig:depot_configuration_variation}
    \end{figure}

    The depot-configuration study evaluates robustness under different depot placements while keeping the customer distribution, demand pattern, fleet size, and remaining generation parameters fixed. The \textit{Eccentric} configuration places the depot near the boundary of the service region, whereas \textit{Random} selects the depot location uniformly at random. The resulting performance is reported in Figure~\ref{fig:depot_configuration_variation}.

    Figure~\ref{fig:depot_configuration_variation} exhibits trends consistent with the previous robustness analyses. In both depot configurations, the Qwen3 variant exhibits the lowest routing distance while maintaining a zero final miss rate. In contrast, CCBC continues to obtain lower raw distances at the expense of substantially higher missed demand. Overall, the proposed adaptive decomposition process yields robust performance under varying depot configurations while preserving high service levels.
    
    Overall, the results suggest that decomposition quality is strongly influenced by the statistical aspects of the instance, making the selection of an appropriate partitioning strategy a critical component of high-volume systems. Fixed approaches optimize a predetermined objective or follow a predefined policy, whereas the proposed approach continuously adapts its decisions according to the evolving state. This allows the framework to balance competing objectives without committing to a single fixed strategy, reliably exhibiting robust routing performance across diverse problem scales and operating conditions.

\section{Conclusion}
    This paper presented an adaptive cluster-first route-second algorithm for large-scale CVRP. By representing decomposition as a sequence of decisions over a hierarchical cluster tree, our approach enables iterative selection of clustering and refinement actions according to the patterns of the evolving routing state, rather than relying on a fixed and proxy objective or predetermined policy. Experimental results on synthetic and benchmark-derived instances demonstrate strong performance on benchmark-scale data while exhibiting strong scalability to substantially larger instances. Throughout diverse routing environments, the method maintains a favorable balance between customer service levels and cost, demonstrating the potential of adaptive, LLM-guided decision making as a practical approach for scalable decomposition and industrial logistics planning.

\section*{Data Availability Statement}
The data supporting the findings of this study are available from the corresponding author, O.K., upon reasonable request.

\section*{Author Contributions Statement}
Oguzhan Karaahmetoglu: Conceptualization, Methodology, Software, Validation, Formal analysis, Investigation, Data curation, Visualization, Writing – original draft.

Hyong Kim: Supervision, Conceptualization, Methodology, Validation, Writing – review \& editing, Project administration.

\section*{Disclosure Statement}
The authors report there are no competing interests to declare.

\bibliography{refs}{}
\bibliographystyle{plain}

\appendix
\section{Decomposition Agent: Splitting Tools}
    \label{appendix:split-tool}
    \noindent The following tool is used to split a certain leaf in the tree.
    
    \begin{lstlisting}[basicstyle=\ttfamily\small, breaklines=true]
    [MUTATE]
    
    Split a partition into child partitions using one customer split method and one vehicle split method.
    
    Customer Methods:
    - uniform: good for spatially uniform distribution of customers
    - kmeans: good for clustered / mixture-like regions
    - sweep: good for uniform / radial regions
    - cap_k: good for balancing demand across child clusters
    - largest_axis: good for clean binary geometric cuts
    
    Vehicle Methods:
    - round_robin_demand: fastest feasibility-oriented balancing
    - equal_count: simple equal vehicle split
    - min_cost_quota: more geometry-aware balancing
    - rebalance: balance vehicles based on capacity/demand gap.
    
    Levels:
    - low: coarse local refinement
    - medium: moderate repartition
    - high: aggressive repartition
    - finalized: estimate the number of child clusters from min split size, aiming for routing-ready leaves
    \end{lstlisting}

\section{Decomposition Agent: Redoing Tools}
    \label{appendix:redo-tool}
    \noindent The following tool is used to redo a partition from the parent level.
    
    \begin{verbatim}
    [MUTATE]
    
    Rebuild an existing subtree under the same parent using a different split strategy.
    
    Direction:
    - merge -> halve the current number of leaf children under this parent
    - split -> double the current number of leaf children under this parent
    
    This is useful when a previous split was too coarse or too fragmented.
    
    Guard:
    - Only parents who
    \end{verbatim}

\section{Decomposition Agent: Stop Tool}
    \label{appendix:stop-tool}
    \noindent The following tool is used to signal the end of decomposition.
    
    \begin{lstlisting}[basicstyle=\ttfamily\small, breaklines=true]
    [STOP]
    
    Stop the clustering process.
    
    Use when the partition tree looks stable enough for routing and further splits are not worthwhile.
    \end{lstlisting}

\section{Decomposition Agent: Analysis Tools}
    \label{appendix:analysis-tools}
    \noindent The following alerts summarize structural issues in the current partition state and guide action selection.
    
    \begin{lstlisting}[basicstyle=\ttfamily\small, breaklines=true]
    [ALERT:max_clusters]
    At/over max partitions. Do NOT split demand nodes further.
    Prefer VEHICLE rebalance or REPAIR (merge).
    \end{lstlisting}
    
    \begin{lstlisting}[basicstyle=\ttfamily\small, breaklines=true]
    [ALERT:split_limit / split_budget]
    Demand node split limit or budget reached/low.
    Do NOT call demand node split tools; prefer VEHICLE rebalance or MERGE.
    \end{lstlisting}
    
    \begin{lstlisting}[basicstyle=\ttfamily\small, breaklines=true]
    [ALERT:size_imbalance]
    Partition size imbalance (max/min ratio too high).
    Prefer splitting the largest partition (not necessarily index 0).
    \end{lstlisting}
    
    \begin{lstlisting}[basicstyle=\ttfamily\small, breaklines=true]
    [ALERT:tiny_parts]
    Very small partitions detected.
    Avoid splitting; consider merging or moving vehicles instead.
    \end{lstlisting}
    
    \begin{lstlisting}[basicstyle=\ttfamily\small, breaklines=true]
    [ALERT:repeat_split]
    Same partition split repeatedly.
    Choose a different target partition.
    \end{lstlisting}
    
    \begin{lstlisting}[basicstyle=\ttfamily\small, breaklines=true]
    [ALERT:pressure]
    Capacity imbalance.
    Under-cap (gap > 0): move/add vehicles.
    Over-cap (gap < 0): donate vehicles or merge.
    \end{lstlisting}
    
    \begin{lstlisting}[basicstyle=\ttfamily\small, breaklines=true]
    [ALERT:vehicle_imbalance]
    Vehicle count imbalance across partitions.
    Prefer VEHICLE rebalance instead of further splitting.
    \end{lstlisting}
    
    \begin{lstlisting}[basicstyle=\ttfamily\small, breaklines=true]
    [ALERT:merge]
    Partitions likely to route poorly (tiny, no vehicles, or vehicle-starved).
    Prefer merging with nearby compatible partitions.
    \end{lstlisting}
    
    \begin{lstlisting}[basicstyle=\ttfamily\small, breaklines=true]
    [CONVERGENCE]
    Checks on max size, size ratio, and capacity gap.
    If satisfied: no major issues, prefer repair-only actions.
    Otherwise: indicates required corrective action (split or rebalance).
    \end{lstlisting}

\section{Decomposition Agent: System Prompt}
    \label{appendix:sysprompt}
    This section provides the system prompt used for the decomposition stage described in Section~III. The prompt defines the decision interface, available actions, and constraints under which the language model operates. The model is not allowed to directly execute actions, but instead proposes structured suggestions that are validated and applied by the system.
    
    \begin{lstlisting}[basicstyle=\ttfamily\small, breaklines=true]
    You are a hierarchical clustering chooser for a routing problem.
    
    Your job is to repeatedly choose EXACTLY ONE mutate tool call.
    You will receive a text analysis of the current active leaves and prioritized alerts.
    
    Main objective:
    - Build a good partition tree quickly for downstream routing.
    - Prefer as few calls as possible.
    - Focus first on leaves highlighted by the alerts, especially the largest and most problematic ones.
    - Prefer broad progress over getting stuck refining one branch too deeply.
    - Do not exceed tree depth 4.
    
    Tree logic:
    - The partitioning is hierarchical. Earlier decisions shape all later descendants.
    - Shallow leaves represent large regions and are usually better places for impactful decisions.
    - Deeper leaves are local refinements and should be used only when the parent region already looks reasonable.
    - If many leaves are still too large, prefer operations that create strong breadth without pushing too deep.
    
    Customer split method guidance:
    - Use largest_axis when the region suggests a clean binary spatial cut.
    - Use kmeans for clustered / multi-center regions.
    - Use sweep for broad radial / angular / coverage-style regions.
    - Use cap_k when balancing demand across child clusters matters.
    - Use uniform for broad, spatially even regions.
    
    Vehicle split method guidance:
    - Use round_robin_demand when speed and fast feasibility matter most.
    - Use equal_count when a very simple even vehicle split is enough.
    - Use min_cost_quota when geometry-aware vehicle assignment is likely to help route quality.
    - Use rebalance when customer groups look uneven in demand / capacity and vehicle capacity should track group demand more closely.
    
    Redo guidance:
    - Use redo_partition when a whole subtree was split with the wrong customer method or wrong aggressiveness.
    - Prefer redo_partition on the parent over repeatedly mutating several child leaves one by one.
    - Use redo to correct tree shape, not as a default action.
    
    Alert priorities:
    - Treat BIG alerts as highest priority when the listed leaves are still far above routing-friendly size.
    - Treat STARVING and VEHICLE LIGHT alerts as signals that vehicle split choice matters more.
    - When alerts are truncated, assume the shown leaves are the most important ones.
    - Prefer acting on the largest shown leaves first unless a strong recovery signal suggests otherwise.
    - If a deep leaf is problematic but its parent subtree looks systematically wrong, act on the parent instead.
    
    Depth guidance:
    - Root and shallow levels should usually use broader structural actions.
    - Depth 4 is the maximum allowed depth.
    - Do not choose actions that would push a branch beyond depth 4.
    - Near the depth limit, prefer better method choice or redo at the parent instead of deeper refinement.
    
    Important:
    - Do not ask questions.
    - Do not explain.
    - Return exactly ONE tool call only.
    \end{lstlisting}

\end{document}